# LukeNet: A lightweight CNN integrated with an XAI model for Smart acute lymphoblastic leukemia detection and management

**Md Taimur Ahad (Corresponding Author)**

Department of Management Information Systems

North South University, Bangladesh

taimur.ahad@northsouth.edu

# LukeNet: A lightweight CNN integrated with an XAI model for Smart acute lymphoblastic leukemia detection and management

***Abstract* —** *Acute Lymphoblastic Leukemia (ALL) is a life-threatening hematological malignancy. ALL patients require early, accurate detection to enable timely treatment and effective patient management. A Convolutional Neural Network (CNN) is well-suited for creating an end-to-end enabling environment for ALL detection and classification. However, most CNN-based ALL detection systems are theoretical and unsuitable for deployment on edge devices due to high computational demands, the need for large datasets, and long training times. ALL patients often exhibit fever, fatigue, rapid heartbeat, and shortness of breath. The Internet of Medical Things (IoMT)-enabled devices offer an opportunity to monitor ALL patients in real time. Wearables that track temperature, heart rate, oxygen saturation, and activity can deliver critical data to support timely clinical intervention and improve patient outcomes. In smart IoMT environments, a lightweight CNN is essential because connected devices often operate under limited computational power, memory, and latency constraints. To address this need, this study proposes LukeNet, a lightweight CNN integrated with explainable artificial intelligence (XAI) for an IoMT-based SMART Acute Lymphoblastic Leukemia Detection and Management System. Trained on three (3) ALL datasets and five-fold cross-validation, LukeNet achieved an impressive 99% model accuracy as well as 99% unseen test accuracy, which is higher than six state-of-the-art (SOTA) CNNs, such as DenseNet121, MobileNet, ResNet50, InceptionV3, Xception, and VGG16, as well as transfer learning models. Furthermore, LukeNet was compared with two ensemble models. In addition, explainable AI methods such as LIME, SHAP, and Grad-CAM are integrated to highlight relevant regions in microscopic images, thereby improving the transparency of predictions. The novelty of this study lies in the architecture of LukeNet, which balances model depth and computational efficiency by using depthwise separable convolutions, mitigates the risk of gradient loss in deeper layers, and provides strong global and local feature extraction capabilities.*

**Keywords:** Acute lymphoblastic leukemia (ALL) detection, Convolutional neural network-based cancer detection, Peripheral Blood Smear, CNN, Deep Learning, Transfer Learning Model, Ensemble model.

# 1. Introduction

Acute lymphoblastic leukemia (ALL) is marked by a high incidence and mortality and by the rapid expansion of malignant lymphoblasts. The World Health Organization reports that the risk of developing ALL is highest in children younger than 5 years of age, but most deaths from ALL occur in adults (Aby et al., 2024; Iqbal, 2018). The rapid transformation of benign cells into malignant cells is a pressing concern for medical providers, as malignant cells can invade surrounding tissues and spread to other parts of the body (Boutry et al., 2022). Furthermore, ALL can aggressively spread to the central nervous system, liver, spleen, and other vital organs, causing organ damage or failure.

The initial screening test for ALL, Peripheral blood smear (PBS), is (Sun, 2022). PBS is a blood sample on a glass slide that, once dried, is stained with May-Grünwald Giemsa. Cancer sizes can be measured precisely using PBS imaging (Ayadi, 2021). Traditional diagnostic methods for detecting ALL using PBS are labor-intensive and error-prone, as human interpretation can vary (Ayadi, 2021; Kibriya, 2022). The detection and classification of ALL can differ significantly based on several factors, including size, shape, contrast, location, texture, and morphological characteristics (Sun, 2022; Deepak, 2019).

Convolutional Neural Networks (CNNs) provide end-to-end ALL detection and classification. CNN's capability is highlighted by its ability to handle intricate patterns and variations in ALL microscopic images. Higher accuracy, shorter diagnosis time, and accurate grade and modality classification are among the advantages of CNNs in ALL detection. CNN variants, such as transfer learning (TL), and ensemble methods have also shown significant promise for detecting ALL (Al-Qazzaz, 2021; Ahad et al., 2021). Scholars such as Arkapravo and Mausumi (2022), Amjad et al. (2021), Saeed et al. (2024), Harem et al. (2022), Hosseini et al. (2023), Rahman et al. (2023), and Wadhah et al. (2021) have applied CNN in ALL detection.

Cancer patients often present with neurological manifestations such as headaches, seizures, cognitive impairments, and behavioral abnormalities, all of which require timely and appropriate clinical attention (Ishfaq et al., 2025). The critical nature of these symptoms underscores the importance of rapid intervention and structured clinical management . In such a scenario, the concept of Medical-of-Things-based SMART cancer detection and management systems is a pressing technological need. The medical domain also emphasizes the seriousness of symptoms and the importance of timely intervention (Tisch Brain Tumor Center, 2025). Prior studies have also raised broader concerns regarding clinical translation. Specifically, Sailunaz et al. (2024), Khalighi et al. (2024), Díaz Pernas et al. (2024), Rasool and Bhat (2023), and Hikmah et al. (2024) observed that although CNNs often achieve high accuracy in laboratory settings, their adoption in clinical workflows remains limited because of insufficient deployment infrastructure and weak integration with hospital IT systems. Earlier, Zech et al. (2018) showed that CNNs trained on X-rays from one hospital may fail to replicate their performance on data from other hospitals, highlighting a lack of generalizability across clinical settings. In a similar vein, Sathitratanacheewin and Pongpirul (2018) and Roberts et al. (2020), and more recently Obeid et al. (2024) and Roberts et al. (2025), emphasized the gap between technical performance and practical usability in CNN-based CAD. Collectively, these findings indicate that progress in Leukemia CAD requires not only accuracy but also lightweight deployability, interpretability, and integration into smart healthcare environments.

This practical requirement creates a direct link between lightweight CNN design and smart healthcare deployment. In such settings, a Lightweight Convolutional Neural Network is methodologically important because inference must be performed on edge devices with limited memory and computational capacity. Thus, the shift from CNN design to IoMT-based monitoring is not incidental: a lightweight model is a necessary foundation for low-latency, low-power, and accessible diagnostic support in smart Leukemia management systems (SCMS), especially for remote doctors, healthcare providers, and rural patients (Ishfaq et al., 2025). The main reported advantages of SCMS are continuous, real-time tracking of a patient's vital signs and tumor markers, enabling early detection and timely intervention. It also supports remote healthcare, personalized treatment, and data-driven insights, improving patient outcomes while reducing hospital visits and overall healthcare costs.

To provide intelligent healthcare solutions, SLMS leverages high-speed internet, advanced sensors, radio frequency identification (RFID), wireless sensor networks (WSNs), and edge computing devices. In these systems, Lightweight Convolutional Neural Networks (CNNs) are essential because inference must be performed on edge devices with limited computational capacity and memory.

Recent advancements in lightweight CNNs integrated with XAI have redefined histopathological imaging, though in oncology, precise pattern recognition is important for cancer detection. The challenge of integrating CNN into Smart Acute Lymphoblastic Leukemia Detection and Management systems (SLMS) is impeding their clinical adoption, despite their laboratory successes.
The use of lightweight architectures enables low-latency, energy-efficient, and accessible diagnostics, particularly benefiting remote practitioners and patients in rural areas (Ishfaq et al., 2025). SCMS supports continuous, real-time tracking of vital signs and tumor biomarkers, early intervention, remote healthcare provision, and personalized treatment strategies. These capabilities enhance patient outcomes while reducing hospital dependence and overall healthcare costs.

The core problem of this study is two-fold. Firstly, there is a lack of a CNN that requires minimal image processing but provides high accuracy. Deepak (2019) also argued that variations across ALL modalities pose substantial challenges for CNNs, as microscopic blood cells exhibit complex intra- and inter-class variations. CNN often lacks the ability to capture correlations among granulocyte and agranulocyte feature maps. Secondly, the CNN architecture is criticized for including unnecessary information from feature maps, poor generalization, and the risk of gradient loss from input PBS images (Jawahar et al., 2024). Among other research problems, Wang & Zhang (2020) warned of low accuracy and high false-positive rates in CNN-based ALL detection. Accuracy in cancer detection and classification across modalities is a concern, as lower detection accuracy and high false-positive rates will narrow the applicability and acceptability of CNNs in cancer research (Hossain, 2023; Aladhadh, 2022). Identifying the correct type and grade of cancer in the early stages is essential in the treatment plan (Shafique & Tehsin, 2018).

To advance ALL detection and address the limitations mentioned above, this study makes the following contributions:

1. A CNN model that requires minimal image processing and is trained using five-fold cross-

validation on three datasets.

2. There is a risk of gradient loss from the PBS image as the design of CNN allows flows the input feature layer from the previous layer, thus the output feature layer may get little information to conduct accurate classification (Jawahar et al., 2024). Therefore, there is a need for a CNN that can control gradient loss regardless of network depth. This study presents *LukeNet*, which was validated in three microscopic PBS datasets to evaluate model robustness.

3. Six states of the arts CNNs DenseNet121, ResNet50, InceptionV3, Xception, MobileNet, and VGG16 were evaluated under two paradigms: trained from scratch and using TL. These comparisons validated the efficiency and effectiveness of *LukeNet* in contrast to more complex architectures

4. To enhance model transparency, SHAP and Grad-CAM were applied to the predictions of *LukeNet*. SHAP visualizations highlighted feature contributions at the pixel level, while Grad-CAM localized the image regions most influential in the decision-making process. These XAI tools improved interpretability and increased clinical trust in automated predictions.

The rest of the paper is organized as follows. Section 2 provides a literature review; Section 3 outlines the details of the experiments; Section 4 presents the results of the experiments; Section 5 is the discussion; Section 6 is the limitation and future research direction; and Section 7 is the conclusion.

# 2. State of the Art

CNNs are applied to ALL detection using architecture modifications, hyperparameter optimization, and adding or removing layers. Following this path, Anand et al. (2025) applied a customized CNN that included five convolutional blocks. A block consists of 13 convolutional layers and five max pooling layers. The results show that the model achieved the best accuracy and precision of 0.96 and 0.95, respectively, using the Adam Optimizer. Thiriveedhi et al. (2025)

integrated CNNs and XAI to diagnose and interpret acute lymphoblastic leukemia. The ALL-Net achieved an average precision of 99.35%, a recall of 99.33%, and an F1 score of 99.58%. The evaluation metrics are higher than ALL-Net, which outperformed EfficientNet, MobileNetV3, VGG-19, Xception, InceptionV3, ResNet50V2, VGG-16, and NASNetLarge, except for DenseNet201. Abbas et al. (2025) presented a lightweight CNN for ALL classification. The model comprises integrated Squeeze-and-Excitation (SE) within Inverted Residual Blocks. The achieved accuracy was 100 %.

Saeed et al. (2024) presented DeepLeukNet, which achieved 99.61% accuracy. Claro (2020) introduced a CNN architecture in this line to differentiate between ALL, acute myeloid leukemia (AML), and healthy blood slides. The study used 16 datasets comprising 2,415 photos. The proposed model was evaluated and achieved 97.18% accuracy and 97.23% precision. Vogado et al. introduced LeukNet, a CNN designed for accurate leukocyte classification. Data augmentation techniques were applied to expand the training dataset, and cross-validation achieved an accuracy of 98.61%. Cross-dataset validation showed that LeukNet outperformed state-of-the-art techniques, achieving accuracies of 97.04%, 82.46%, and 70.24% across three datasets (Vogado, 2021). Vo et al. (2022) proposed a method using deep learning algorithms and microscopic blood smear images to automatically detect and classify malaria and acute lymphoblastic leukemia (ALL). The method consists of three stages: segmentation with a modified UNet, classification with a convolutional neural network, and data fusion with a perceptron. The proposed approach achieves an overall accuracy of 93%, with a 95% detection rate for ALL and 92% for malaria. This method provides a reliable, interpretable solution for detecting abnormal leukocytes in ALL and identifying malaria-infected blood cells (Vo, 2022). Sampathila et al. developed a deep learning-based CNN to differentiate leukemic cells from normal blood cells. Their customized ALLNET model achieved impressive performance metrics, including 95.54% accuracy, 95.81% specificity, 95.91% sensitivity, 95.43% F1-score, and 96% precision. (Sampathila, 2022).

# 3. Research Methodology

In this study, we aim to develop a lightweight CNN, LukeNet, to detect and classify ALL PBS images. Two important criteria for LukeNet are: first, the model must be lightweight and edge-device-integratable, and second, it must achieve high accuracy on unseen datasets. The LukeNet model was tested on three (3) ALL datasets having a varied number of classes and images.

The LukeNet model was trained using K-fold stratified cross-validation. To ensure that each fold trains the same proportion of class labels as the original dataset, K-5 stratified cross-validation was selected. The LukeNet model performance was compared with six (6) prominent CNNs, transfer learning, and two ensemble models. We also conducted computational cost and statistical analysis of LukeNet with SOTA CNNs to confirm its robustness.  The study's workflow is presented in Figure 1. We first describe the hardware and software for the study.

The research methodology describes the resources for the experiments, including the datasets, hardware, programming environment, and model development (see Figure 1).

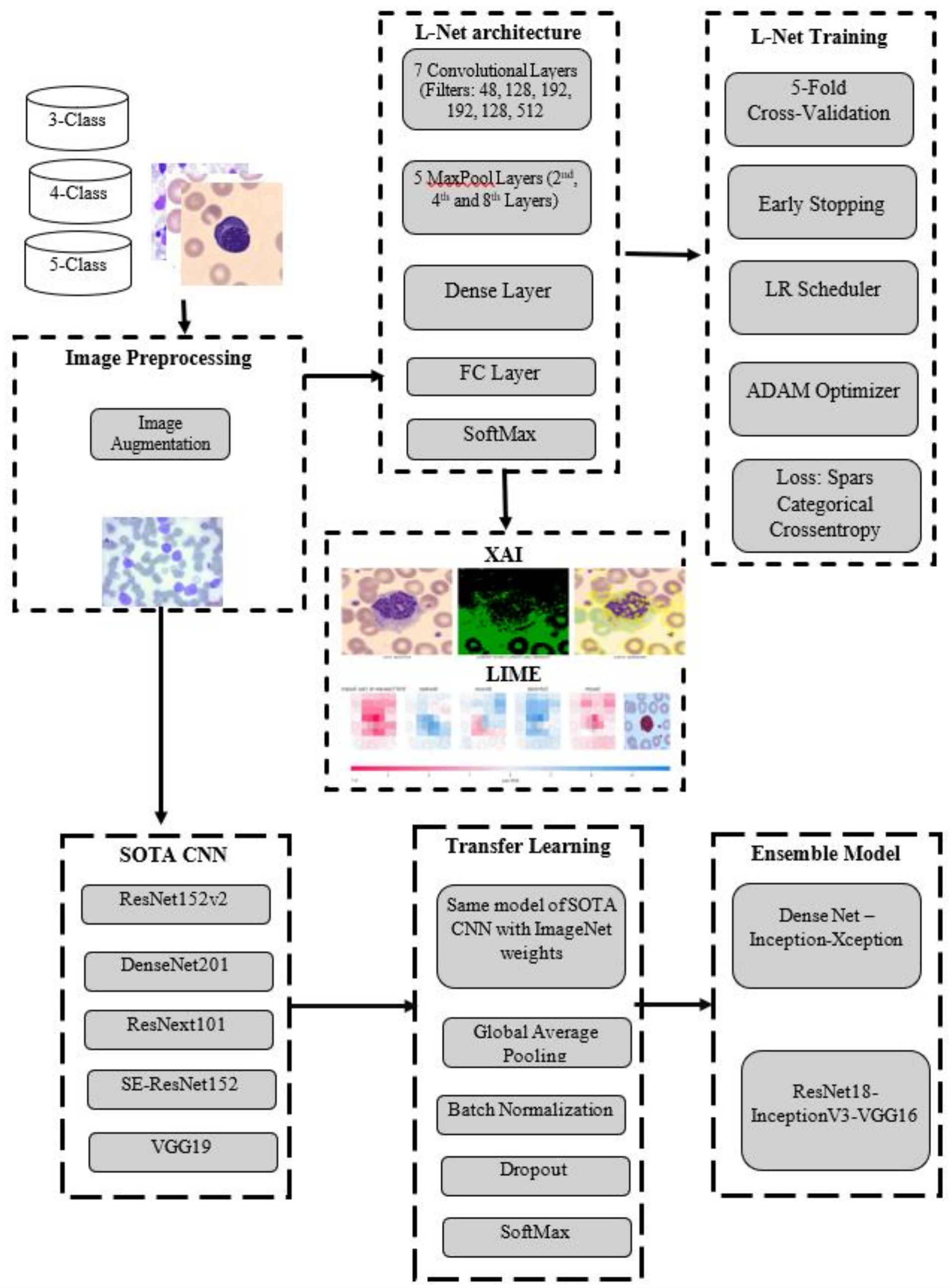

Figure 1: Workflow of the proposed methodology

### 3.1 Experiment Platform and Dataset

The GCN models were developed on a Precision 7680 Workstation with a 13th Gen Intel Core i9-13950HX vPro processor, and an NVIDIA RTX 3500 Ada Generation GPU running Windows 11 Pro. Python was the programming environment for this research. The TensorFlow/Keras framework was selected because it supports Spektral's GCNConv layers for graph convolution operations. TensorFlow/Scikit-learn was also helpful for feature extraction from the histopathology images. The details of the experiment are presented in Figure 1.

In this study, three (3) datasets were used to analyze the performance of the LukeNet. The datasets were collected from publicly available respiratory datasets. Dataset A has four classes and 3242 PBS images. The PBS images were collected from several Tehran hospitals (Iran). The images were obtained from 89 patients suspected of ALL. The dataset includes a benign case and three malignant cases. The images were captured with a Zeiss camera from the optical microscope and saved as JPG files. The link to the dataset is (https://www.kaggle.com/datasets/mohammadamireshraghi/blood-cell-cancer-all-4class).

The Dataset-B consists of 5 classes, including one regular class and four types of leukemia: Acute Lymphoblastic Leukemia (ALL), Acute Myeloid Leukemia (AML), Chronic Lymphocytic Leukemia (CLL), and Chronic Myeloid Leukemia (CML). Each class includes 1000 images in the data set. The link to the data set is (https://www.kaggle.com/datasets/sumithsingh/blood-cell-images-for-cancer-detection).

The Dataset-C includes eight classes and 17,092 ALL images. The normal blood cells include neutrophils, eosinophils, basophils, lymphocytes, monocytes, promyelocytes, myelocytes, metamyelocytes, and thrombocytes. The link to the data set is (https://www.kaggle.com/datasets/mahdinavaei/blood-cancer). The dataset was collected using the CellaVision DM96 analyzer at the Clinical Hospital of Barcelona.

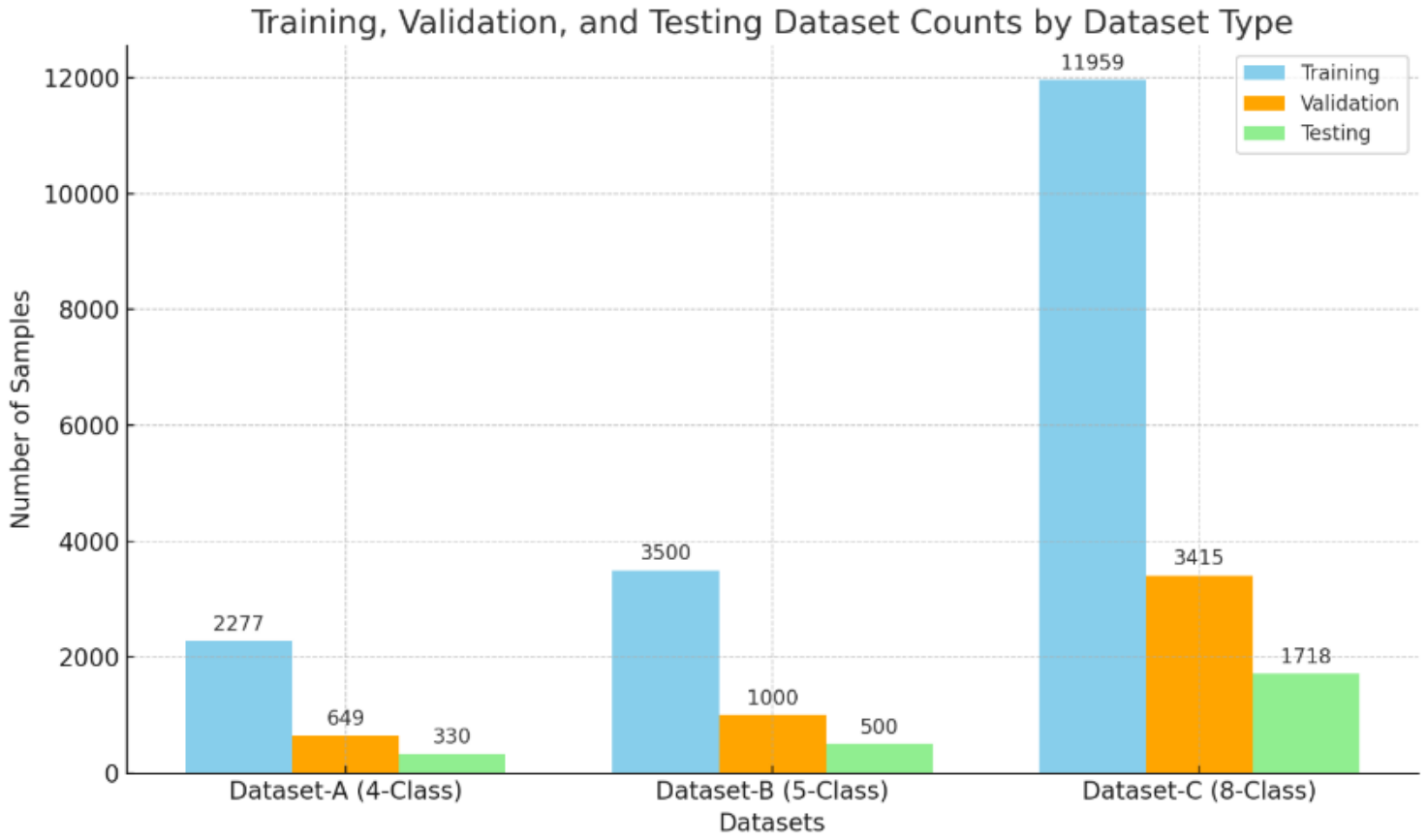

Figure 2: Distribution of three (3) datasets

The datasets are divided according to the 70:20:10 rule. That means 70% of the data was used for training, 20% for validation, and 10% for testing (See Figures 2 and 3).

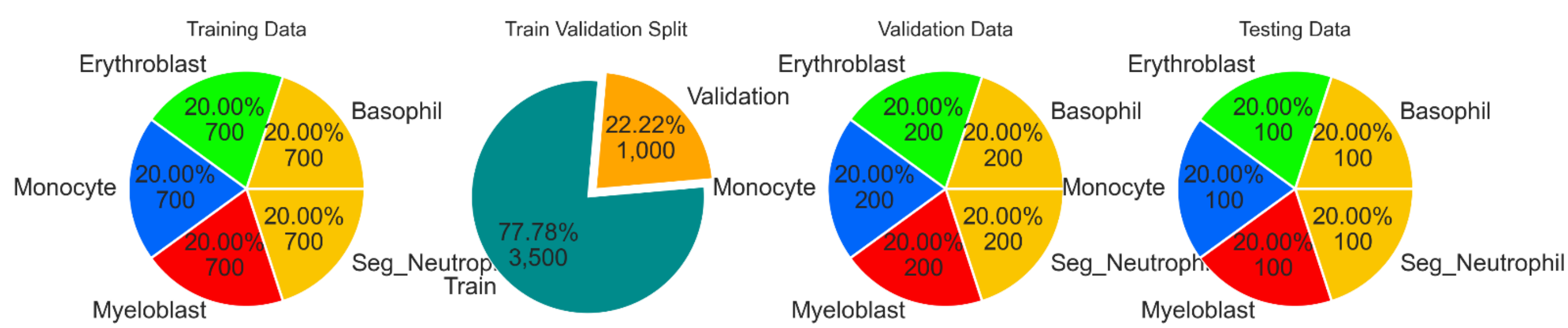

Figure 3: Example of distribution of Dataset-B (5-Class)

The sample PBS image dataset is shown in Figure 4.

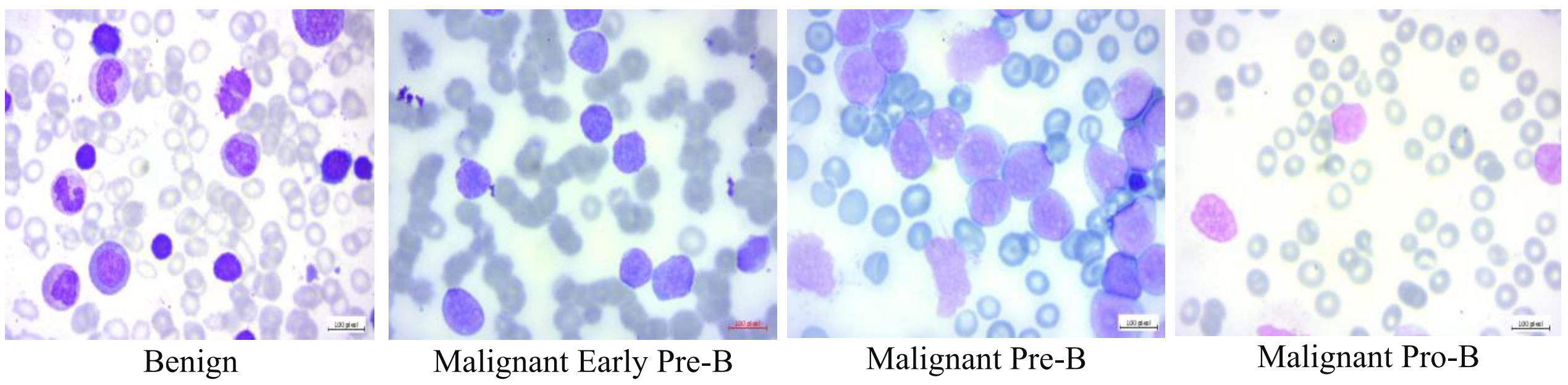

Figure 4: Sample images of Dataset-B (5-Class)

### 3.2 Image augmentation process

Data augmentation techniques are used to increase the number of images in a training dataset. The augmentation process improves the performance of CNN models. Therefore, CNN's interest in implementing data augmentation has increased—and so has deep learning research (Nagaraju et al., 2022). Image augmentation is the process of expanding an existing dataset by transforming the original data to create new, label-preserving data (Sankupellay Konovalov, 2018; Meeras et al., 2019).

Image augmentation was performed using scaling, cropping, flipping, rotation, and color adjustments, including brightness, contrast, and saturation. The data augmentation process involved various transformations, including random rotations within the range of -15 to 15 degrees, rotations at multiples of 90 degrees, random distortion, shear, vertical and horizontal flipping, skewing, and intensity transformation. Augmented image samples are shown in Figure 5.

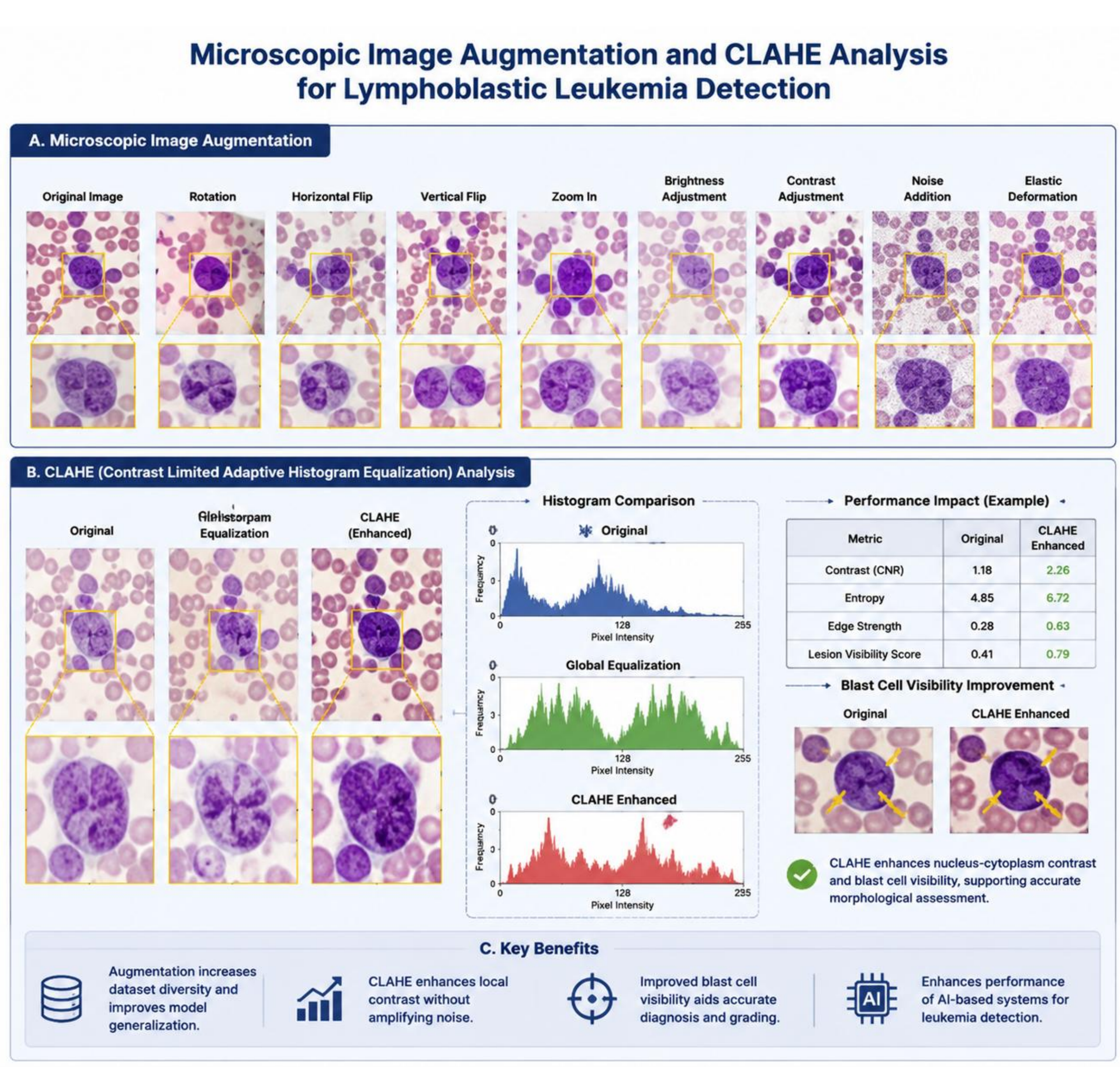

Figure 5: Sample augmented images

### 3.3 LukeNet model description

The model consists of 6 convolutional layers, one depth-wise separable convolutional layer, global average pooling, and dense layers with regularization and dropout (see Table 1). The layers are organized in blocks. The convolutional layers progressively increase in depth (32, 64, 128, 256, 512, 1024 filters) within each block. The progressive convolution captures features from simple to complex in the PBS image. The architecture adds additional blocks with more filters to make the network deeper, which is useful for extracting more abstract, higher-level features from the image.

The inclusion of a depth-wise separable convolution splits the filtering and channel combination into two separate operations. Therefore, this convolution is computationally efficient and reduces the number of parameters compared to the standard Conv2D.

After each convolutional layer, a batch normalization layer is applied to normalize the layer's outputs. Batch Normalization reduces internal covariate shift, stabilizes training, and accelerates training. Batch Normalization layers are applied after each convolutional operation to normalize activations, improving training stability and reducing overfitting. Batch Normalization normalizes the activations across the batch to improve training stability. Normalization helps prevent issues like internal covariate shift (where the distribution of activations changes during training).

Dropout is also applied between dense layers at high rates (0.45 and 0.4), further helping reduce overfitting by randomly setting some neurons to zero during training. Dropout is applied to prevent overfitting by randomly setting a fraction p of the input units to 0 during training. For example, the dropout rate is set to 0.45 for the first dense layer and 0.4 for the second dense layer.

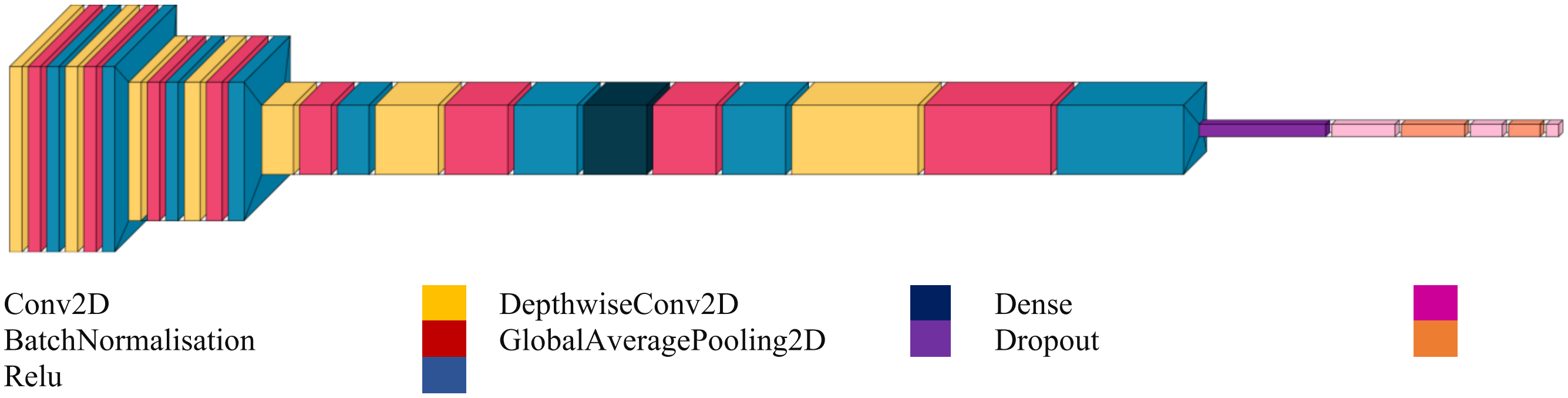

Figure 6: A visual structure of the LukeNet model

Table 1. Architecture details of the proposed LukeNet model. Summary of each layer, including type, output shape, and parameter count

| Task | Layer Name | Input Shape | Output Shape | Parameters |
|---|---|---|---|---|
| Feature Extraction | Conv2D | (112, 112, 3) | (112, 112, 32) | 896 |
| | BatchNormalization | (112, 112, 32) | (112, 112, 32) | 128 |
| | ReLU | (112, 112, 32) | (112, 112, 32) | 0 |
| | Conv2D | (112, 112, 32) | (112, 112, 64) | 18,496 |
| | BatchNormalization | (112, 112, 64) | (112, 112, 64) | 256 |
| | ReLU | (112, 112, 64) | (112, 112, 64) | 0 |
| | Conv2D | (112, 112, 64) | (56, 56, 128) | 73,856 |
| | BatchNormalization | (56, 56, 128) | (56, 56, 128) | 512 |
| | ReLU | (56, 56, 128) | (56, 56, 128) | 0 |
| | Conv2D | (56, 56, 128) | (56, 56, 256) | 295,168 |
| | BatchNormalization | (56, 56, 256) | (56, 56, 256) | 1,024 |
| | ReLU | (56, 56, 256) | (56, 56, 256) | 0 |
| | Conv2D | (56, 56, 256) | (28, 28, 512) | 1,180,160 |
| | BatchNormalization | (28, 28, 512) | (28, 28, 512) | 2,048 |
| | ReLU | (28, 28, 512) | (28, 28, 512) | 0 |
| | Conv2D | (28, 28, 512) | (28, 28, 1024) | 4,719,616 |
| | BatchNormalization | (28, 28, 1024) | (28, 28, 1024) | 4,096 |
| | ReLU | (28, 28, 1024) | (28, 28, 1024) | 0 |
| | DepthwiseConv2D | (28, 28, 1024) | (28, 28, 1024) | 10,240 |
| | BatchNormalization | (28, 28, 1024) | (28, 28, 1024) | 4,096 |
| | ReLU | (28, 28, 1024) | (28, 28, 1024) | 0 |
| | Conv2D | (28, 28, 1024) | (28, 28, 2048) | 2,099,200 |
| | BatchNormalization | (28, 28, 2048) | (28, 28, 2048) | 8,192 |
| | ReLU | (28, 28, 2048) | (28, 28, 2048) | 0 |
| | GlobalAveragePooling | (28, 28, 2048) | (2048) | 0 |
| Classification | Dense | (2048) | (1024) | 2,098,176 |
| Regularization | Dropout | (1024) | (1024) | 0 |
| Classification | Dense | (1024) | (512) | 524,800 |
| Regularization | Dropout | (512) | (512) | 0 |
| Classification | Dense | (512) | (4) | 2,052 |

Instead of using flattening or global max pooling, this model uses Global Average Pooling (GAP) to reduce the spatial dimensions and create a more compact representation before the fully connected layers. This method reduces overfitting and improves generalization. In the mode, the bias regularizer is applied to the fully connected layers (Dense) to regularize the bias terms. The model uses Regularization techniques, such as L2 and L1, in the dense layers to prevent overfitting, which is particularly useful when the model is deep and complex. L1 bias regularization is applied to the biases of the dense layers with a regularization strength of 0.006.

Since there is a significant increase in the depth of the fully connected layers with 1024 and 512 neurons, combining these layers with regularization techniques such as L1 and L2 regularization and dropout helps the network learn more complex patterns from the convolutional feature maps. The output layer uses the SoftMax activation function to produce a probability distribution over the possible classes, which is standard for multi-class classification.

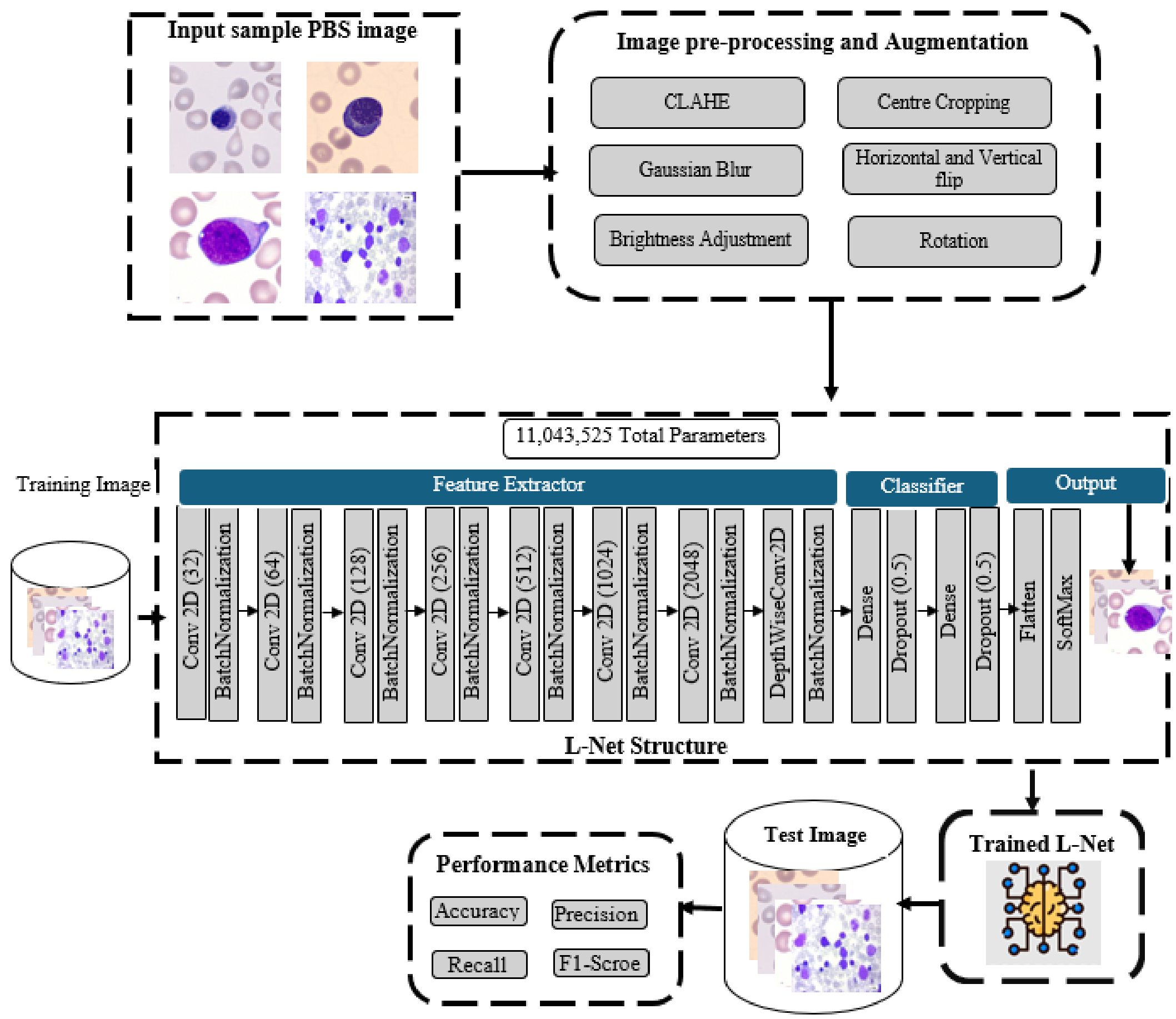

Figure 7: Layered architecture of LukeNet.

---

Algorithm 1 Training LukeNet with XAI integration

---

Require: Image dataset D, LukeNet model architecture.
Ensure: Best-performing LukeNet model with explainable outputs.

```
1    Initialize LukeNet using build_LukeNet_model()
2    for fold k =1 to 5 do
3        Split D into train_data, val_data using split_dataset(D, k)
4        Reset model weights
5        Compile   model with  Adam(learning_rate = 1e-4) and SparseCategoricalCrossentropy()
6      for epoch e=1 to 50 do
7          for each batch (x,y) in train_data do
8              Compute predictions using the model(x)
9              Calculate loss and apply gradients via GradientTape()
```

| | |
|---|---|
| 10 | end for |
| 11 | end for |
| 12 | Evaluate model on val_data using model.evaluate() |
| 13 | Save validation accuracy for the current fold. |
| 14 | Generate XAI outputs on validation samples: |
| 15 | SHAP: DeepExplainer (model, background_data).shap_values() |
| 16 | Grad-CAM: GradCAM. generate_heatmap() |
| 17 | Store all explanations using save_explanations() |
| 18 | end for |
| 19 | Compute average validation accuracy across all folds. |

## 3.4 LukeNet training process

The LukeNet training process includes a series of convolutional layers, batch normalization layers, activation functions, pooling layers, dropout layers, and fully connected layers. The input image is first fed into the first Convolutional Layer. The first convolution layer includes 32 Filters, (3X3) Kernel size, (2X2) Strides for down-sampling the spatial dimensions using the same Padding. If the input image is X, W is the kernel (filter) tensor, and b is the bias, then the convolution can be expressed as:

$$\mathrm{Conv1}(X) = \mathrm{ReLU}(\mathrm{Conv2D}(X, W_1) + b_1)$$

This convolution extracts low-level features, such as edges and textures, and the resulting features pass through a Batch Normalization layer. If μ is the mean, σ2 is the variance, and ϵ is a small constant to prevent division by zero, the output of $\mathrm{Conv1}(X)$ is normalized using

$$x^{=\sigma 2} + \epsilon x - \mu$$

ReLU is the activation function applied after batch normalization. For the input $x$, the $ReLU(x)$ can be expressed as:

$$ReLU(x) = \max(0, x)$$

The normalised feature map is then used as input to the second convolutional layer. The deeper the layer, the more complex shapes the model captures. However, in the second convolution by Conv2, captures the outlines of blood cells, the curve of tissue structures and or other meaningful regions in the PBS image detected by Conv1:

$$\mathrm{Conv2}(X) = \mathrm{ReLU}(\mathrm{Conv2D}(\mathrm{Pool1}(X), W_2) + b_2)$$

The output Y of input X can be expressed as

$$Y = Conv2D(X, W) + b\ x^{=\sigma 2} + \epsilon x - \mu$$

In block 6, a Depthwise Separable Convolution reduces the number of parameters by separating the convolution into two stages, compared to a standard convolutional layer. First, each input channel is convolved with its own filter using depth-wise convolution. Second, a pointwise convolution: a 1x1 convolution to mix the output channels. This is mathematically represented as: $Y_{\text{dw}} = \text{DepthwiseConv}(X)$ followed by $Y_{\text{pw}} = \text{Conv2D}(Y_{\text{dw}}, W_{\text{pw}}) + b_{\text{pw}}$

After the convolutional and depthwise separable layers, the model performs Global Average Pooling (GAP), which computes the average of all spatial locations for each channel:

$$y_i = \frac{1}{H \times W} \sum_{h=1}^{H} \sum_{w=1}^{W} x_{i,h,w}$$

Here, H and W are the height and width of the feature map, $x_{i,h,w}$ is the value at the $(h, w) - th$ Location in the i-th feature map, $y_i$ is the output of the global average pooling operation for the i-th feature map. This produces a fixed-length vector that serves as input to the fully connected layers.

Following the Global Average Pooling (GAP) layer, the model has two dense layers with regularization. The first dense layer has 1024 units, with L2 kernel regularization and L1 regularization for the activity and bias. If W and b are the weight matrix and bias vector, respectively, the dense layer output for a given input $X_{\text{gap}}$ is calculated as:

$$Y_{\text{dense}} = WX_{\text{gap}} + b$$

The regularizes used are L1 and L2, which help prevent overfitting by penalizing large weights or large activations. The L1 regularization term is:

$$\mathcal{L}_1 = \lambda \sum_i |W_i|$$

Dropout is applied to the Dense layer's output to prevent overfitting. The dropout rate is set to 0.45 for the first dense layer. If X is the input to the dropout layer, then the output of the dropout layer is:

$$Y_{\text{dropout}} = \text{Dropout}(X)$$

The second dense layer has 512 units with similar regularizations. If L2 is the regularisation term and λ is the regularisation coefficient, and $W_i$ are the weights being regularised then the L2 can be mathematically expressed as:

$$\mathcal{L}_2 = \lambda \sum_i W_i^2$$

The second Dense layer uses a dropout rate of 0.4. The output of the dropout layer is:

The final layer is a dense layer with SoftMax activation. This layer outputs a probability distribution across C classes, where each class corresponds to a specific category in the classification task. For an input $X_{fc}$ The output is:

$$y_c = \frac{e^{z_c}}{\sum_{j=1}^{C} e^{z_j}}$$

Where $e^{z_c}$ is the output of the dense layer, and $y_c$ is the probability for class c, the SoftMax can be expressed as below:

$$f(X) = \text{Softmax}\left(\text{Dense}\left(\text{Dropout}\left(\text{Dense}\left(\text{GAP}\left(\text{DepthwiseConv}\left(\text{Conv}\left(\text{ReLU}\left(\text{BN}\left(\text{Conv}(X)\right)\right)\right)\right)\right)\right)\right)\right)\right)$$

### 3.5 Training hypermeters

The parameters used for training and validating the LukeNet model are exhibited in Table 2.

Table 2. Hyperparameter tuning for LukeNet

| Parameter | Value |
|---|---|
| Epochs | 50 |
| BatchSize | 32 |
| ImageSize | 64,64,3 |
| LearningRate | 1.0000e-04 |
| WeightDecay | 0.0000001 |
| K-folds | 5 |
| Optimizer | Adam(learningrate=1.0000e-04) |
| Loss | SparseCategoricalCrossentropy(from_logits=True) |
| EarlyStopping | Monitor:val_accuracy,patience=10,restorebestweights |
| LearningRateScheduler | 0.1×learningrateevery10epochs |
| Callbacks | EarlyStopping, LearningRateScheduler |

# 4. Results of experiments

The experiments' findings are presented in three sections, focusing on the original, transfer learning, and ensemble methodologies.

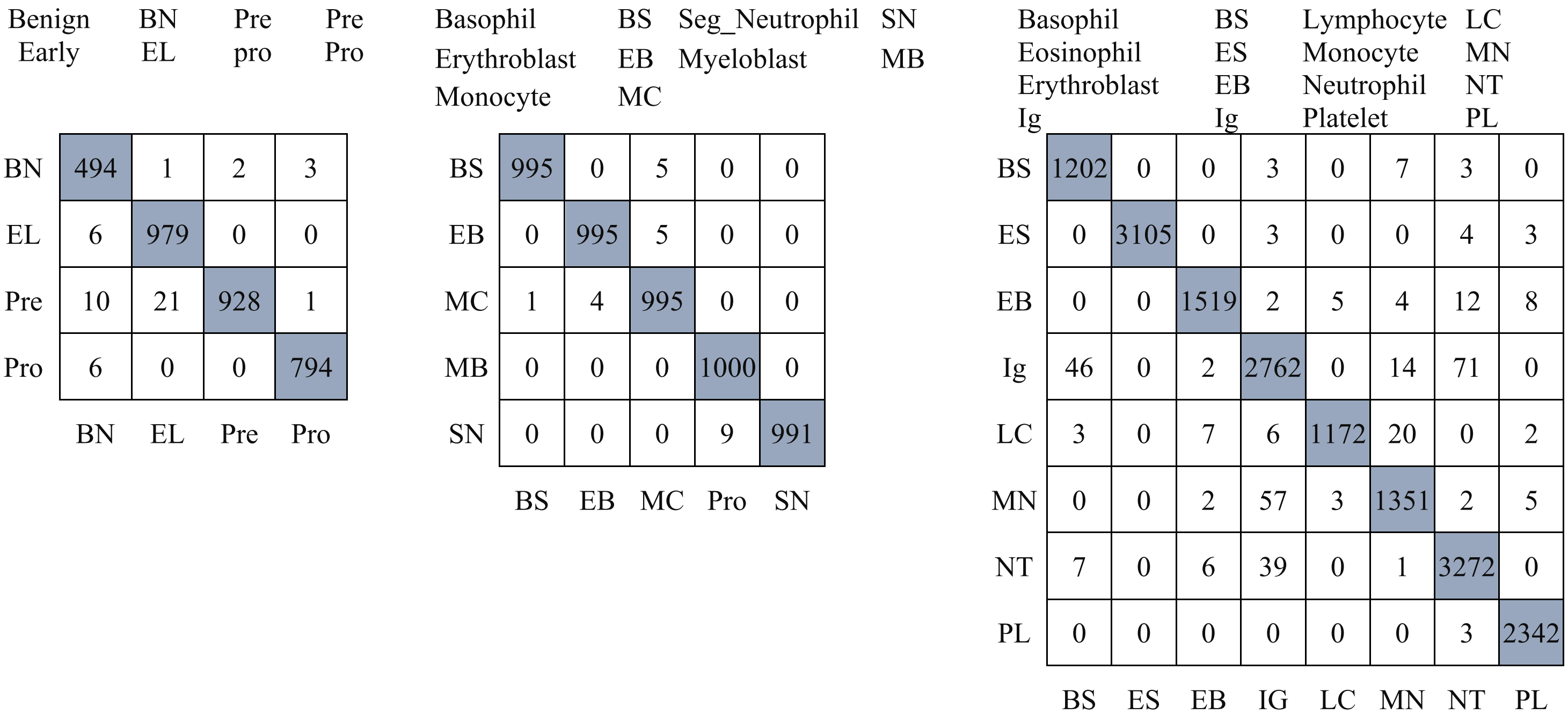

Figure 8: Confusion matrix of three datasets of the LukeNet model

Figure 8 displays the confusion matrix of three datasets. LukeNet's Early (EL) has 979 correctly classified, with the others misclassified as Benign (BN), Pre (Pre), or Pro (Pro). Pre (Pre) has 928 correctly classified instances, with a few misclassified as Benign (BN), Early (EL), or Pro (Pro). Pro (Pro) has 794 correctly classified, with errors in the other three classes. In Dataset-B (5-Class), the model shows minimal Type 1 errors. In Dataset-C (8-Class), Type 1 errors are also relatively few. Basophil (BS) has 3 instances misclassified as Eosinophil (ES), and Eosinophil (ES) has 4 instances misclassified as Basophil (BS). Erythroblast (EB) has 2 instances misclassified as Monocyte (MN), while Ig (Ig) has 46 instances misclassified into other categories. Lymphocyte (LC) has 20 misclassified instances, Monocyte (MN) has 57 instances misclassified, Neutrophil (NT) has 6, and Platelet (PL) has 3 instances misclassified as other classes.

## 4.1 Classification Report

Table 3. Classification report of the LukeNet model.

|  | Class | Precision | Recall | F1-Score | Support |
|---|---|---|---|---|---|
| **Dataset-A (4Class)** | Benign | 96% | 99% | 97% | 500 |
| | Early | 98% | 99% | 99% | 985 |
| | Pre | 100% | 97% | 98% | 960 |
| | Pro | 99% | 99% | 99% | 800 |
| | **Accuracy** | | | 98% | 3245 |
| | **Macro Avg** | 98% | 99% | 98% | 3245 |
| | **Weighted Avg** | 98% | 98% | 98% | 3245 |
| **Dataset-B (5-Class)** | **Class** | **Precision** | **Recall** | **F1-Score** | **Support** |
| | Basophil | 100% | 99% | 100% | 1000 |
| | Erythroblast | 100% | 99% | 100% | 1000 |
| | Monocyte | 99% | 99% | 99% | 1000 |
| | Myeloblast | 99% | 100% | 100% | 1000 |
| | Seg_Neutrophil | 100% | 99% | 100% | 1000 |
| | **Accuracy** | | | 100% | 5000 |
| | **Macro Avg** | 100% | 100% | 100% | 5000 |
| | **Weighted Avg** | 100% | 100% | 100% | 5000 |
| **Dataset-C (8-Class)** | **Class** | **Precision** | **Recall** | **F1-Score** | **Support** |
| | Basophil | 96% | 99% | 97% | 1215 |
| | Eosinophil | 100% | 100% | 100% | 3115 |
| | Erythroblast | 99% | 98% | 98% | 1550 |
| | Ig | 96% | 95% | 96% | 2895 |
| | Lymphocyte | 99% | 97% | 98% | 1210 |
| | Monocyte | 97% | 95% | 96% | 1420 |
| | Neutrophil | 97% | 98% | 98% | 3325 |
| | Platelet | 99% | 100% | 100% | 2345 |
| | **Accuracy** | | | 98% | 17075 |
| | **Macro Avg** | 98% | 98% | 98% | 17075 |
| | **Weighted Avg** | 98% | 98% | 98% | 17075 |

The classification of LukeNet is presented in Table 3. The report suggests that in Dataset-A (4-Class), the model achieves 98% accuracy, with high precision (0.96–1.00) and recall (0.97–0.99). Dataset B (5-class) exhibits better performance, with an accuracy of 100%. Similarly, in Dataset-C (8-Class), the LukeNet achieved an accuracy of 98%, and high precision, recall, and F1-scores across most classes. However, only for "Basophil" and "Ig", the model achieves 96% accuracy, with some misclassifications.

## 4.2 Training accuracy and Data loss curve

Table 4. Accuracy and data loss classification report of LukeNet

| | Training Accuracy | Validation accuracy | Test accuracy | Training Data Loss | Validation data loss | Test data loss |
|---|---|---|---|---|---|---|
| Dataset-A | 99% | 99% | 99% | 32% | 31% | 31% |
| Dataset-B | 99% | 99% | 99% | 27% | 28% | 28% |
| Dataset-C | 99% | 99% | 99% | 25% | 24% | 24% |

The training/validation accuracy and loss over time are presented in Table 4. In the case of Dataset-A (4-Class), the training accuracy increases rapidly and reaches a high value early in the epoch. However, the validation accuracy is somewhat lower and increases more gradually, suggesting some overfitting during the initial epoch. The training loss decreases sharply, while the validation loss follows a similar trend but remains slightly higher, suggesting some degree of overfitting. However, overfitting is observed, as the validation loss and accuracy lag the training metrics.

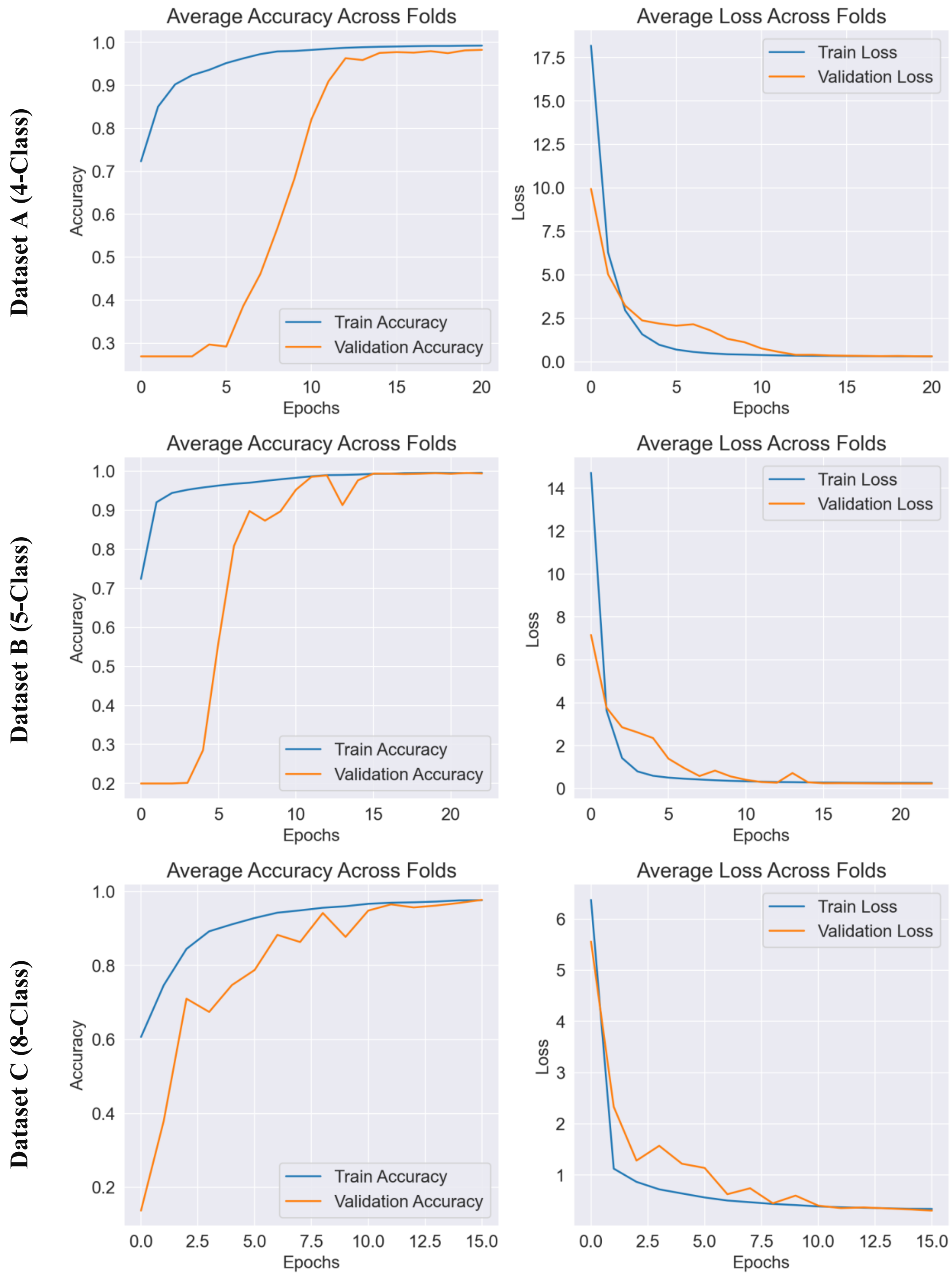

Figure 9: Average accuracy and data loss curve of LukeNet

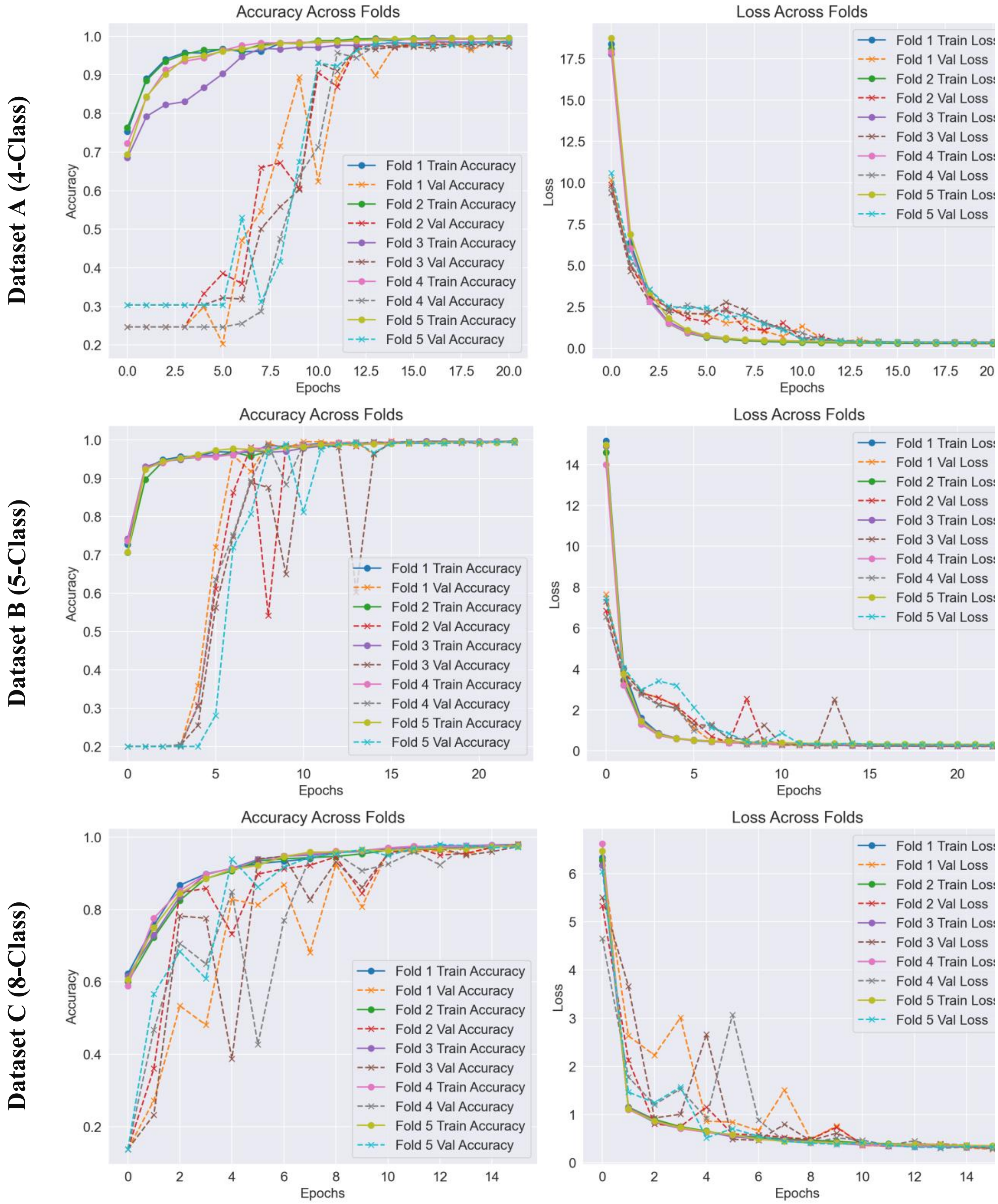

Figure 10: Accuracy and data loss curve of all folds

## 4.3 Benchmark comparison

Table 5. Classification report of SOTA CNNs

Dataset A (4-Class)

**Original CNN**

| Model | Training Accuracy | Validation Accuracy | Test Accuracy |
|---|---|---|---|
| DenseNet121 | 99% | 81% | 94% |
| ResNet50 | 99% | 76% | 93% |
| InceptionV3 | 99% | 81% | 94% |
| Xception | 99% | 83% | 95% |
| MobileNet | 99% | 74% | 92% |
| VGG16 | 30% | 30% | 30% |

**Transfer Learning**

| Model | Training Accuracy | Validation Accuracy | Test Accuracy |
|---|---|---|---|
| DenseNet121 | 93% | 80% | 89% |
| ResNet50 | 70% | 53% | 64% |
| InceptionV3 | 94% | 79% | 90% |
| Xception | 92% | 76% | 87% |
| MobileNet | 95% | 80% | 90% |
| VGG16 | 86% | 79% | 84% |

Dataset B (5-Class)

**Original CNN**

| Model | Training Accuracy | Validation Accuracy | Test Accuracy |
|---|---|---|---|
| DenseNet121 | 100% | 99% | 99% |
| ResNet50 | 99% | 97% | 99% |
| InceptionV3 | 99% | 98% | 99% |
| Xception | 99% | 98% | 99% |
| MobileNet | 99% | 92% | 97% |
| VGG16 | 20% | 20% | 20% |

**Transfer Learning**

| Model | Training Accuracy | Validation Accuracy | Test Accuracy |
|---|---|---|---|
| DenseNet121 | 94% | 92% | 94% |
| ResNet50 | 63% | 63% | 63% |
| InceptionV3 | 96% | 92% | 95% |
| Xception | 94% | 92% | 93% |
| MobileNet | 96% | 95% | 96% |
| VGG16 | 83% | 82% | 82% |

Dataset C (8-Class)

**Original CNN**

| Model | Training Accuracy | Validation Accuracy | Test Accuracy |
|---|---|---|---|
| DenseNet121 | 99% | 94% | 98% |
| ResNet50 | 100% | 92% | 97% |
| InceptionV3 | 99% | 94% | 98% |
| Xception | 99% | 94% | 98% |
| MobileNet | 99% | 85% | 95% |
| VGG16 | 19% | 19% | 19% |

**Transfer Learning**

| Model | Training Accuracy | Validation Accuracy | Test Accuracy |
|---|---|---|---|
| DenseNet121 | 88% | 81% | 86% |
| ResNet50 | 47% | 42% | 46% |
| InceptionV3 | 89% | 80% | 86% |
| Xception | 84% | 75% | 81% |
| MobileNet | 90% | 82% | 87% |
| VGG16 | 61% | 55% | 59% |

The LukeNet was compared with DenseNet121, MobileNet, ResNet50, InceptionV3, Xception, and VGG16, as well as transfer learning models. The investigation suggests that LukeNet performs more reliably and accurately. However, DenseNet121 and Xception consistently perform well in both original-CNN and transfer-learning scenarios across datasets. Transfer learning generally improved the performance of models like Xception and MobileNet. These models seem to benefit the most from pre-trained weights. VGG16, despite transfer learning, did not show significant improvement, indicating that this model may need further optimisation for transfer learning. The results are shown in Table 5.

## 4.4 Ensemble model

Two ensemble models were developed to evaluate LukeNet's performance. This study's first ensemble model (DVS) is developed by combining multi-path (DenseNet201), spatial exploitation (VGG19), and Feature Map Exploitation (SE-ResNet152). The algorithms presented below:

Algorithm 2: Ensemble procedure.

1: Input: [DenseNet-201, VGG19, SEresNet152], test_dataset
2: Output: ensemble_prediction
3: **models** ← [DenseNet-201, VGG19, SEresNet152]
4: for all *model in models* do
5: **predictions** ← `model_predict` (*test_dataset*)
6: end for
7: **pred_array** ← `array` (*predictions*)
8: **pred_sum** ← `sum` (*pred_array*, *axis* = 0)
9: *ensemble_pred* ← `argmax` (*pred_sum*, *axis* = 1)
10: **ensemble_prediction** ← *ensemble_pred*

Table 6: Training and model accuracy of the Ensemble model (DenseNet201, VGG19, and SEresNet152)

| **Ensemble CNNs** | **Training Accuracy** | **Model Accuracy** |
|---|---|---|
| DenseNet-201, VGG19 and SEresNet152 | 97.44% | 98.76% |

Table 7: Precision, Recall, F1-Score, and Support of Ensemble Model

| **Ensemble** | | | | |
|---|---|---|---|---|
| | Benign | Malignant Early Pre-B | Malignant Pre-B | Malignant Pro-B |
| Precision | 98% | 92% | 95% | 99% |
| Recall | 87% | 95% | 97% | 99% |
| f1-score | 92% | 93% | 96% | 99% |
| support (N) | 1657 | 3206 | 3172 | 2589 |

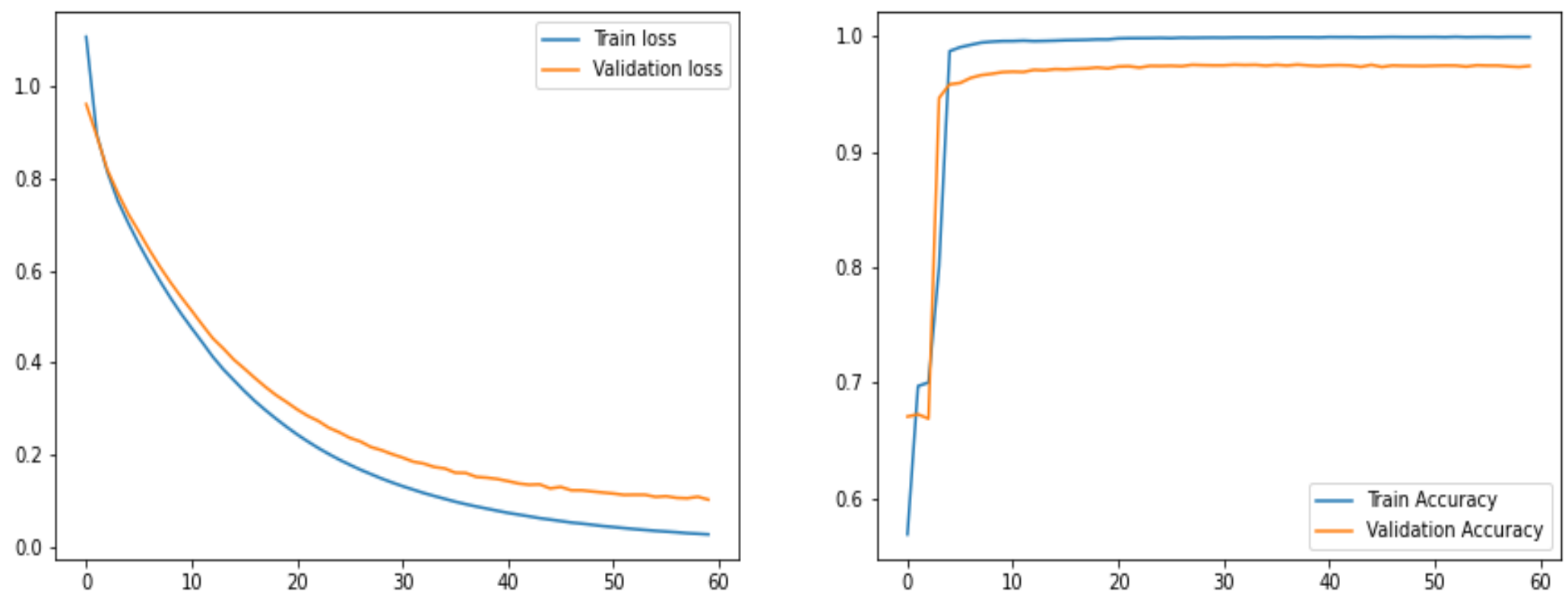

Figure 11: Data loss curve and accuracy curve of the Ensemble model. The Figure shows no overfitting and that the training and validation data are appropriately divided.

DVS performance is shown in Tables 6 and 8. Figure 11 shows the data-loss and accuracy curves for the DVS. The number of epochs is shown on the x-axis, and the accuracy and loss percentages are shown on the y-axis. The Figure shows no overfitting and that the training and validation data are appropriately divided.

## 4.5 Ensemble model-2

In the second ensemble, DenseNet201, InceptionV3, and Xception were selected as the candidates for the ensemble model 'DIX'. The purpose was to combine a strong classifier with a weak classifier to validate the ensemble's capabilities, as suggested by Sharma et al. (2023). The results are presented in Tables 8 and 9 and Figure 12.

Algorithm 1 Ensemble procedure.

1: Input: [DenseNet201, InceptionV3, Xception],
test_dataset 2: Output: ensemble_prediction
3: **models** ← [DenseNet201, InceptionV3,
Xception] 4: for all *model in models* do
5: **predictions** ← `model_predict`
(*test_dataset*) 6: end for
7: **pred_array** ← `array` (*predictions*)
8: **pred_sum** ← `sum` (*pred_array*, *axis* = 0)
9: *ensemble_pred* ← `argmax` (*pred_sum*, *axis* = 1)
10: **ensemble_prediction** ← *ensemble_pred*

Table 8. Training and model accuracy of the Ensemble model DIX

| Architecture | Training Accuracy | Model Accuracy |
|---|---|---|
| DenseNet-201, InceptionV3 and Xception (DIX) | 97.63% | 99.12% |

Table 9. Precision, Recall, F1-Score and Support of Ensemble model DIX

| DIX | | | | |
|---|---|---|---|---|
| | Benign | Malignant Early Pre-B | Malignant Pre-B | Malignant Pro-B |
| precision | 100% | 95% | 100% | 100% |
| Recall | 88% | 100% | 100% | 100% |
| f1-score | 93% | 97% | 99% | 100% |
| support (N) | 1672 | 3254 | 3198 | 2628 |

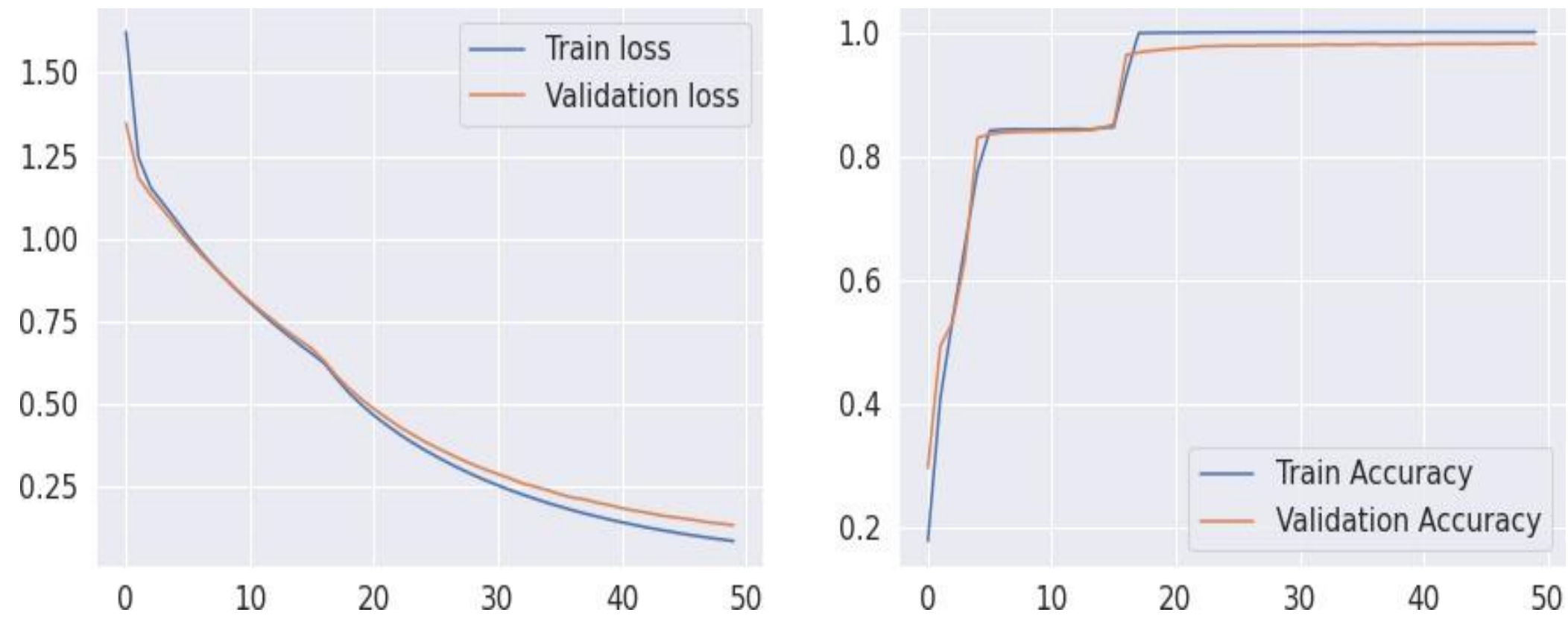

Figure 9: Loss and Accuracy curve of Ensemble model DIX.

The ensemble model outperformed the original CNN architecture, achieving 99.12% accuracy. If any model shows lower accuracy, the ensemble model is used to improve overall accuracy and performance. Additionally, the ensemble DIX model gives 0.83% better performance than the original CNN architecture and 4.12% better performance than transfer learning. Table 8 presents the Precision, Recall, F1-score, and Specificity results of the CNN networks using the ensemble. Figure 9 illustrates the training and validation accuracies of the ensemble model trained on DenseNet201 data. Figure 9 illustrates that there is no overfitting, and the training and validation data are appropriately separated.

### 4.6 Computational cost of the LukeNet and deep learning models

LukeNet demonstrates outstanding efficiency across multiple performance metrics. In terms of memory usage, it maintains the lowest among SOTA models (see Figure 10.A). Its parameter count remains consistent across different classification tasks, making it highly efficient and scalable without increasing complexity.

When evaluating latency, LukeNet also outperforms many SOTA models with significantly lower milliseconds per image (see Figure 10.B). This is particularly beneficial for real-time applications, where responsiveness is key. Despite its lightweight architecture, LukeNet holds up well against models like MobileNet, which exhibit higher latency on more complex tasks.

In terms of GFLOPS, LukeNet is also computationally efficient (see Figure 10.C). While other models, such as MobileNet, show high computational cost, LukeNet requires far fewer floating-point operations, making it suitable for edge devices and environments with limited processing power.

When examining inference memory, LukeNet continues to lead in efficiency (see Figure 10.D). Its memory consumption is significantly lower than that of DenseNet121 and VGG16, making it ideal for deployment in memory-constrained devices. The model's design minimizes resource usage, ensuring smooth operation in various real-world applications.

Lastly, LukeNet's training memory usage remains minimal compared to other SOTA networks (see Figure 10.E). It maintains a consistent memory footprint regardless of task complexity, showcasing its suitability for training large datasets or handling complex tasks without requiring substantial hardware resources. This memory efficiency, combined with its low latency and parameters, makes LukeNet a strong candidate for practical deployment in constrained environments.

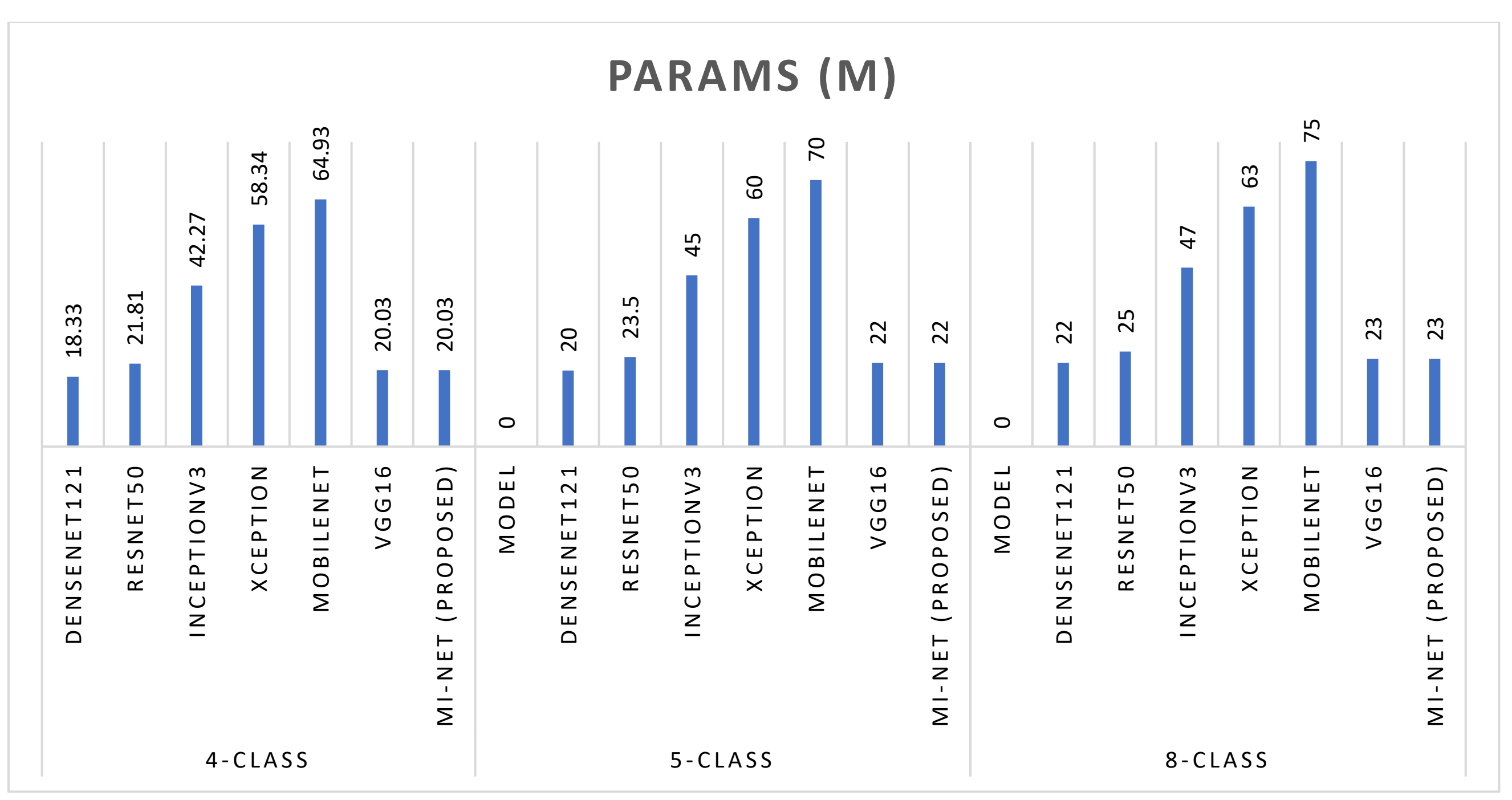

Figure 10 (A): Parameter comparison between LukeNet and another SOTA CNNs

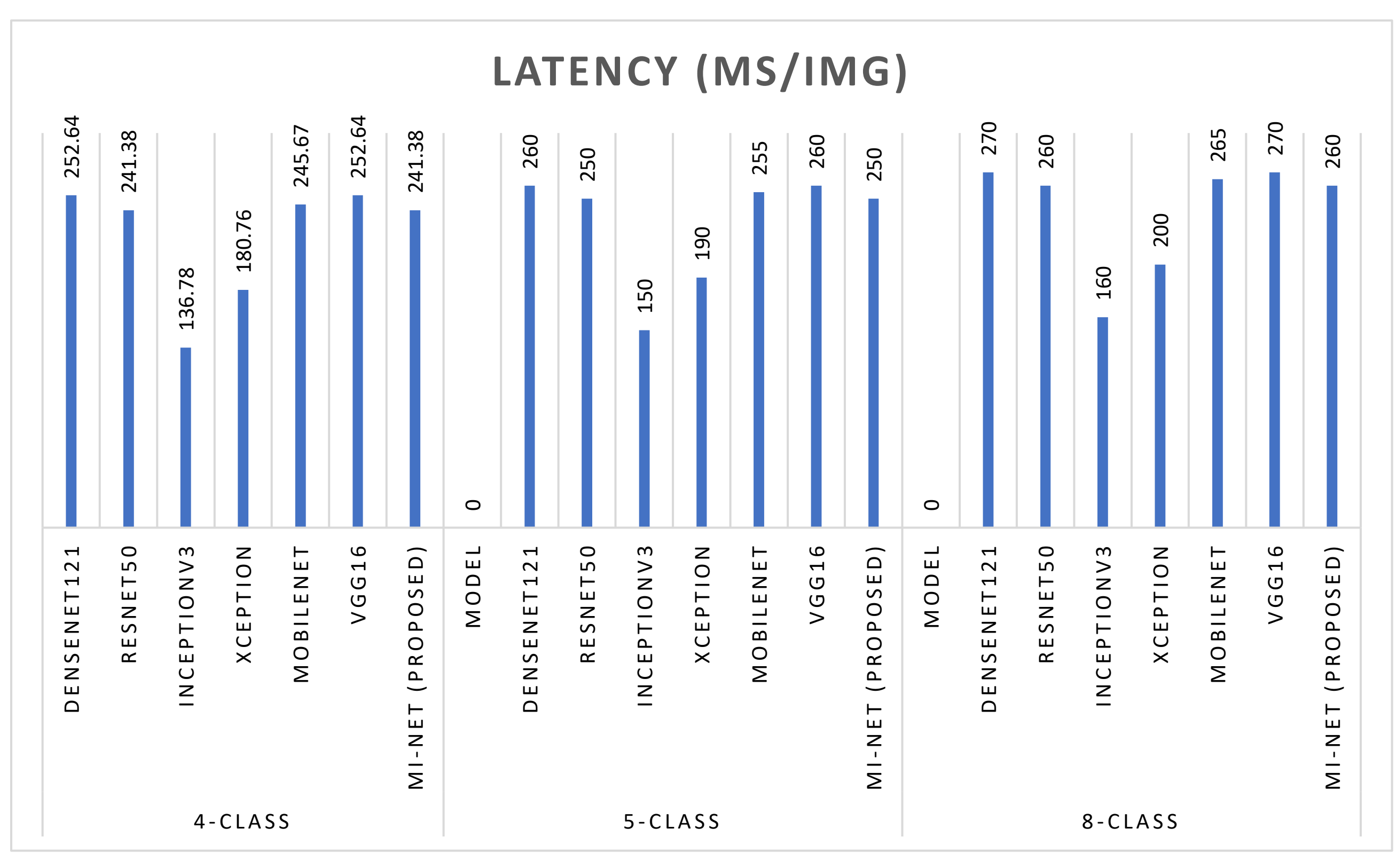

Figure 10 (B): Latency (ms/img) comparison between LukeNet and other SOTA CNNs

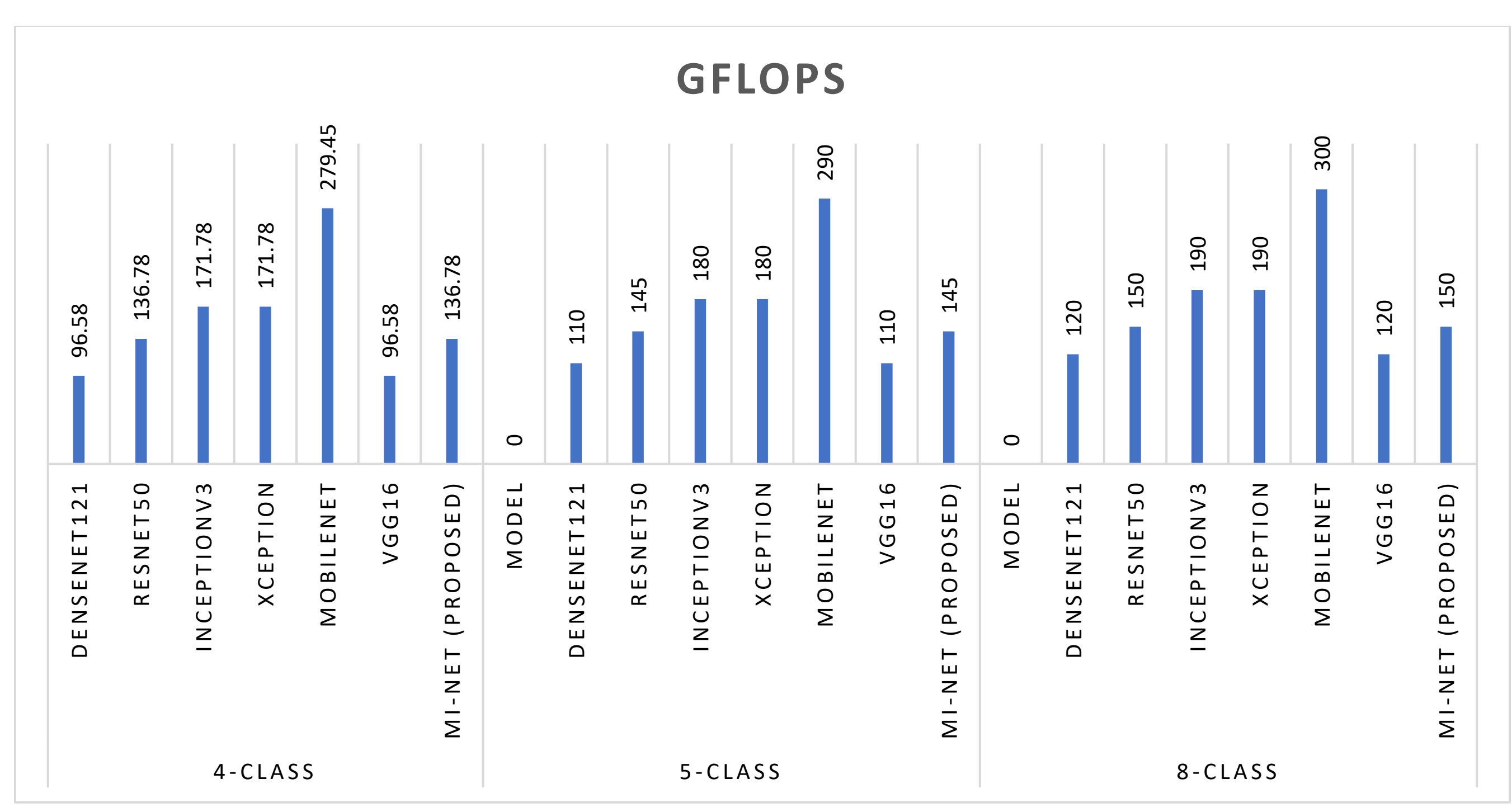

Figure 10 (C): GFOPS comparison between LukeNet and other SOTA CNNs

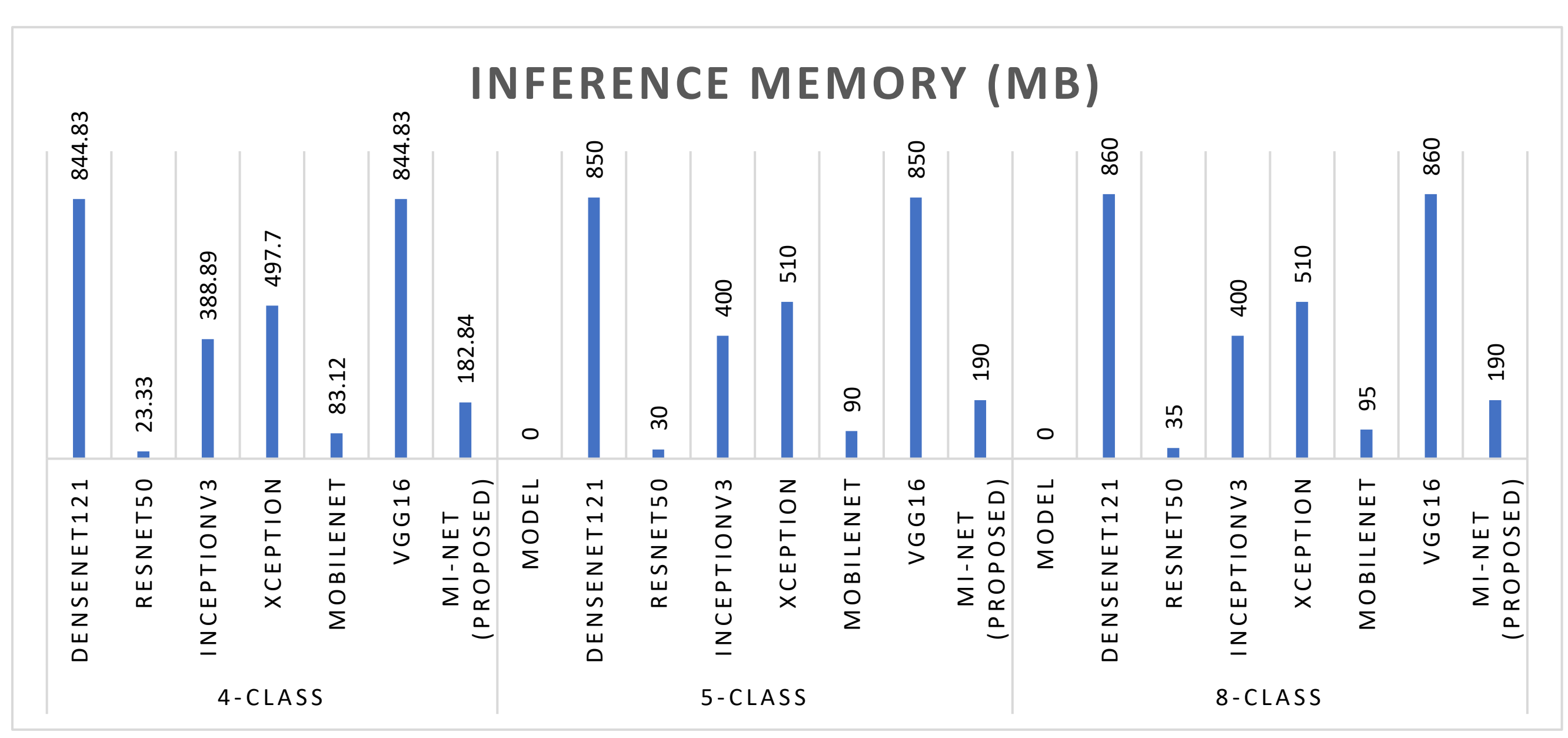

Figure 10 (D): Inference Memory (MB) comparison between LukeNet and other SOTA CNNs

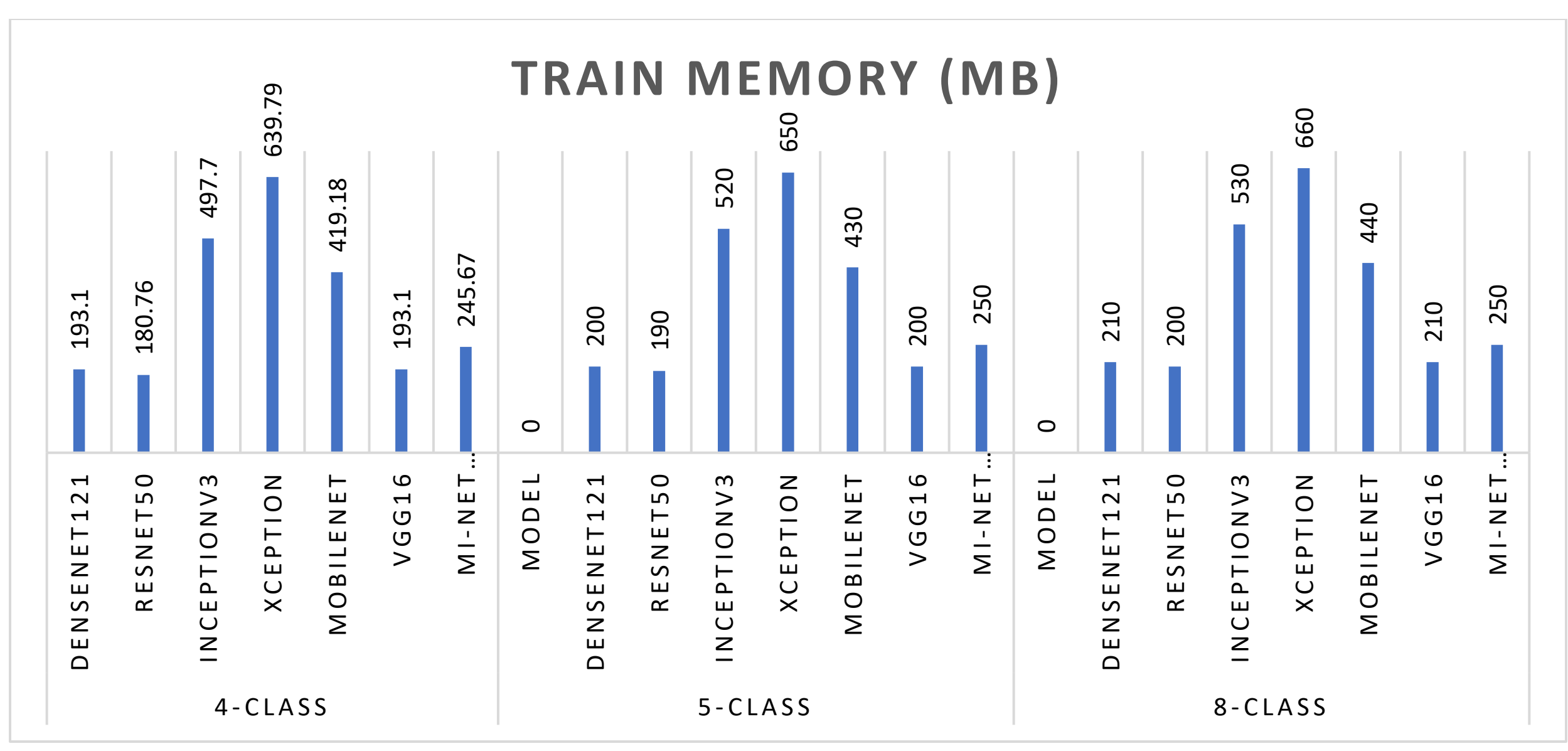

Figure 10 (E): Train Memory (MB) comparison between LukeNet and other SOTA CNNs

## 4.1 Descriptive statistics analysis

Table 8: Statistical analysis of models

| Metric | LukeNet | VGG16 | DenseNet121 | Inception V3 | ResNet50 | Xception | MobileNet |
|---|---|---|---|---|---|---|---|
| Acc Mean | 91.3% | 80.5% | 82.1% | 92.6% | 93.3% | 91.3% | 87.5% |
| Acc Std | 15.4% | 20.4% | 18.7% | 9.4% | 9.0% | 9.7% | 16.7% |
| Acc CI Low | 84.9% | 70.1% | 73.1% | 89.0% | 90.1% | 87.1% | 81.9% |
| Acc CI High | 97.6% | 91.0% | 91.1% | 96.3% | 96.6% | 95.5% | 93.2% |
| Loss Mean | 23.0% | 48.5% | 45.2% | 20.2% | 18.2% | 23.7% | 33.7% |
| Loss Std | 38.5% | 47.8% | 43.5% | 24.4% | 23.4% | 25.3% | 43.7% |
| Loss CI Low | 7.1% | 24.1% | 24.3% | 10.7% | 9.8% | 12.8% | 18.9% |
| Loss CI High | 38.9% | 72.9% | 66.1% | 29.6% | 26.6% | 34.6% | 48.5% |
| Acc Cohen's d | 1.93 | 1.95 | 1.38 | 3.84 | 3.33 | 2.39 | 1.25 |
| Acc Hedges' g | 1.74 | 1.77 | 1.20 | 3.57 | 3.08 | 2.07 | 1.14 |
| Val Acc | 98.9% | 93.8% | 95.6% | 99.1% | 99.5% | 98.6% | 93.1% |
| Test Acc | 98.8% | 93.6% | 96.1% | 99.4% | 99.2% | 98.6% | 90.6% |

The descriptive statistics suggest that LukeNet stands out as a strong performer compared to other SOTA CNNs (see Tables 8 and 9). LukeNet achieves a high mean accuracy of 91%, which is competitive, coming very close to the top performers like ResNet50 and Inception V3. In addition to its strong accuracy, LukeNet exhibits a relatively low standard deviation (15) in accuracy, indicating consistent performance across different runs. The confidence interval for LukeNet's accuracy (85% to 97%) is also relatively tight, showing that its performance is both high and stable.

Table 9: Loss calculation of LukeNet and SOTA CNNs

| Model | Mean | Std | Median | IQR | Range |
| --- | --- | --- | --- | --- | --- |
| DenseNet121 | 36.9% | 29.7% | 24.8% | 27.5% | 98.8% |
| ResNet50 | 185.5% | 244.9% | 76.5% | 145.0% | 870.5% |
| InceptionV3 | 72.9% | 47.0% | 55.8% | 74.9% | 167.2% |
| Xception | 49.5% | 34.8% | 38.0% | 36.0% | 110.7% |
| MobileNet | 73.6% | 52.3% | 59.2% | 72.0% | 170.3% |
| VGG16 | 207.9% | 0.1% | 207.8% | 0.0% | 0.2% |
| LukeNet | 52.6% | 49.3% | 30.5% | 53.3% | 169.1% |

### 4.7 Explainability with SHAP and Grad-CAM

To ensure clinical transparency and trust in LukeNet's predictions, two post-hoc XAI techniques were utilised: SHAP for pixel-level feature attribution and Grad-CAM for region-based visual explanations. SHAP heatmaps (see Figure 11) illustrate the contribution of individual pixels to the final prediction. In subfigure Fig. 9(a), regions marked in blue indicate features that support the non-cancer prediction. In contrast, in subfigure Figure 12, red-highlighted areas correspond to features contributing to the cancer prediction. LIME visualisations demonstrate that LukeNet effectively focuses on diagnostically relevant structures, such as dense masses or irregular textures. A Grad-CAM analysis (Figure 13, 14) was applied to visualise LukeNet's spatial attention during classification. For the correctly classified images, Grad-CAM heatmaps highlighted cancerous areas in positive cases with intense red regions, indicating the model's focus on relevant features. In negative cases, the heatmaps successfully ignored cancerous regions, showing appropriate attention to non-cancerous features. Grad-CAM visual explanations reinforce the model's ability to localise diagnostically significant areas, offering insights into its decision-making process.

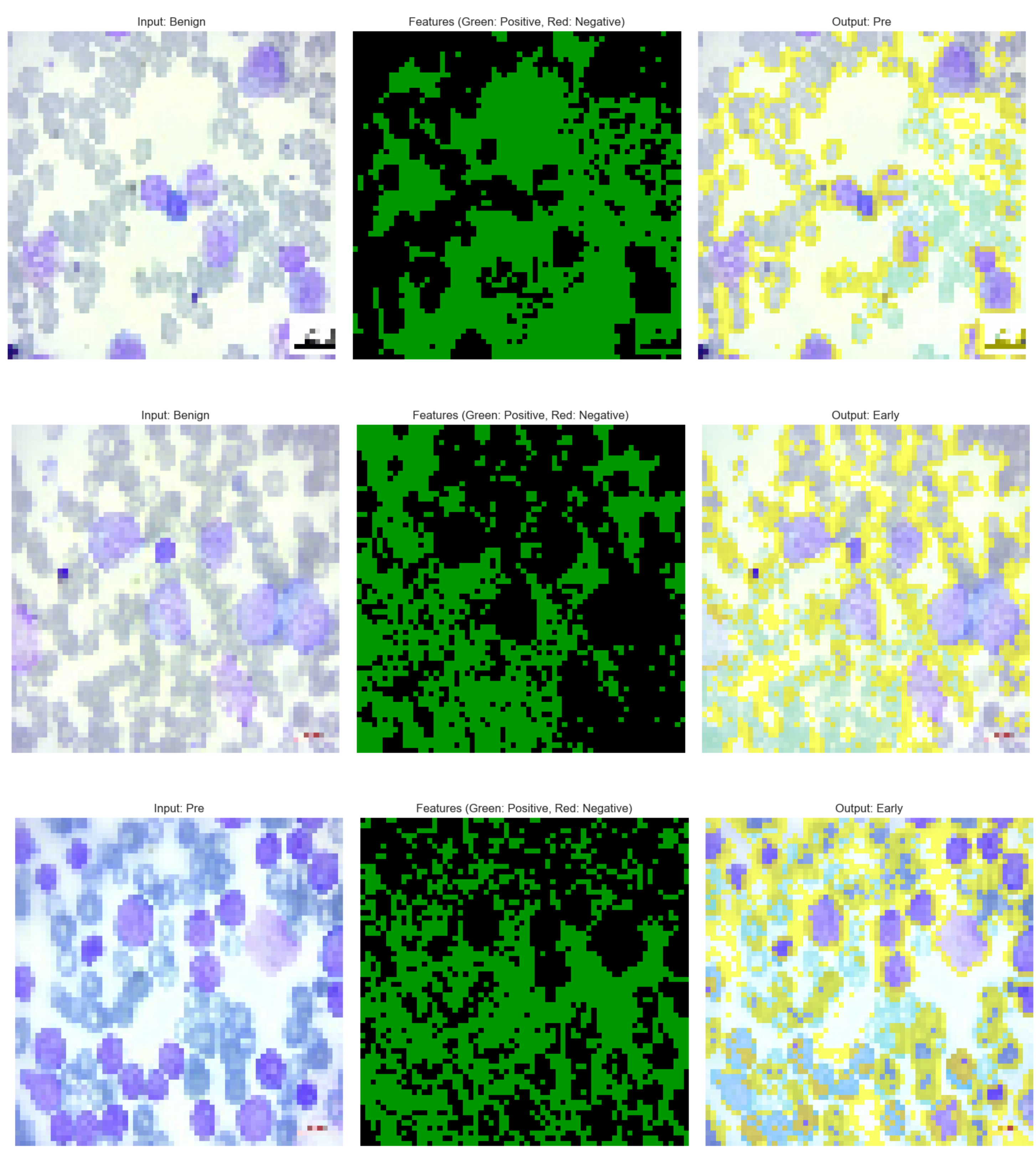

Figure 11: LIME view. Heatmaps illustrating spatial attention for representative correct and incorrect classifications

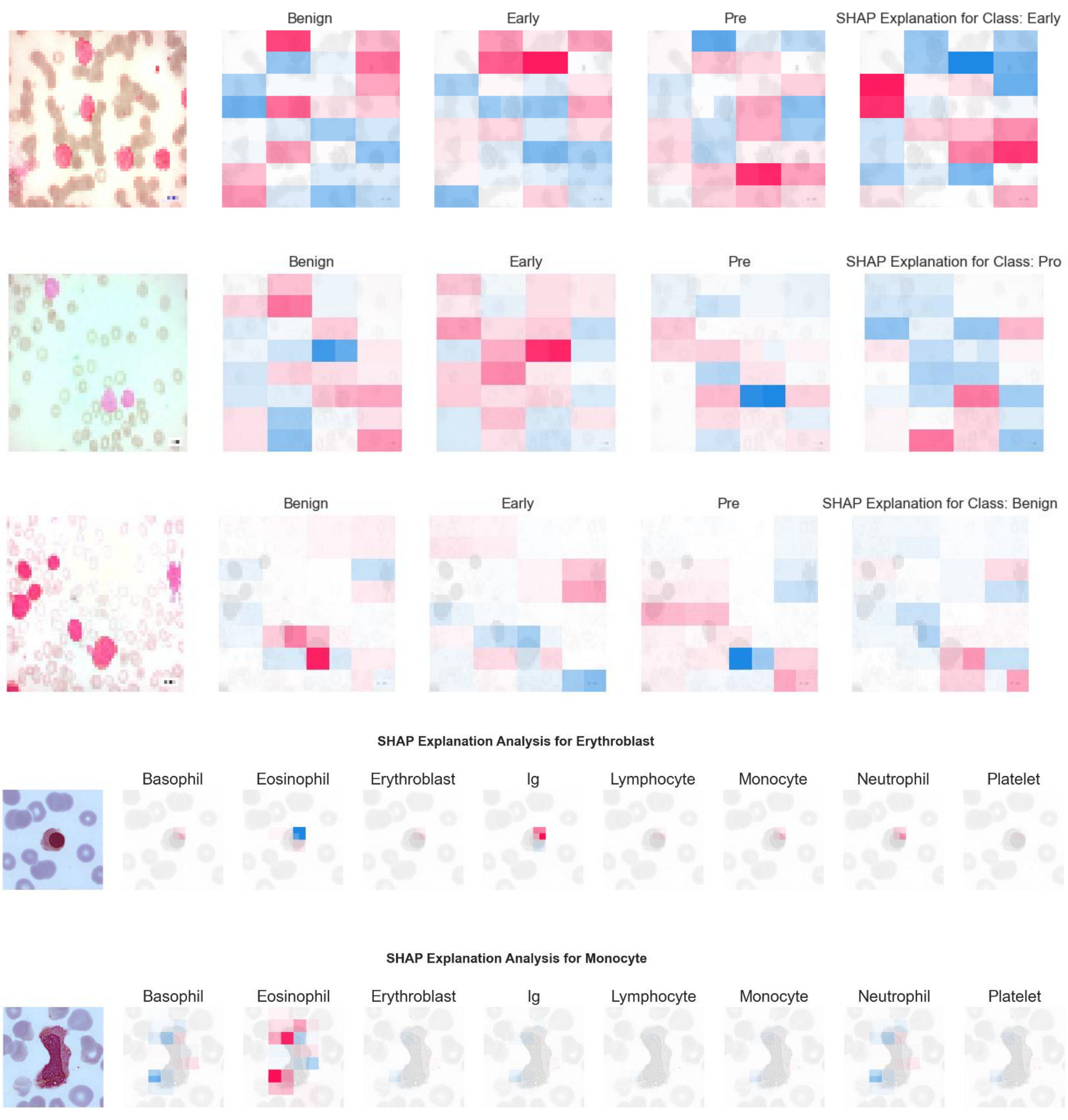

Figure 12: SHAP visualization for detecting the correct prediction of cancer. Heatmaps showing pixel-level contributions to predictions for classes

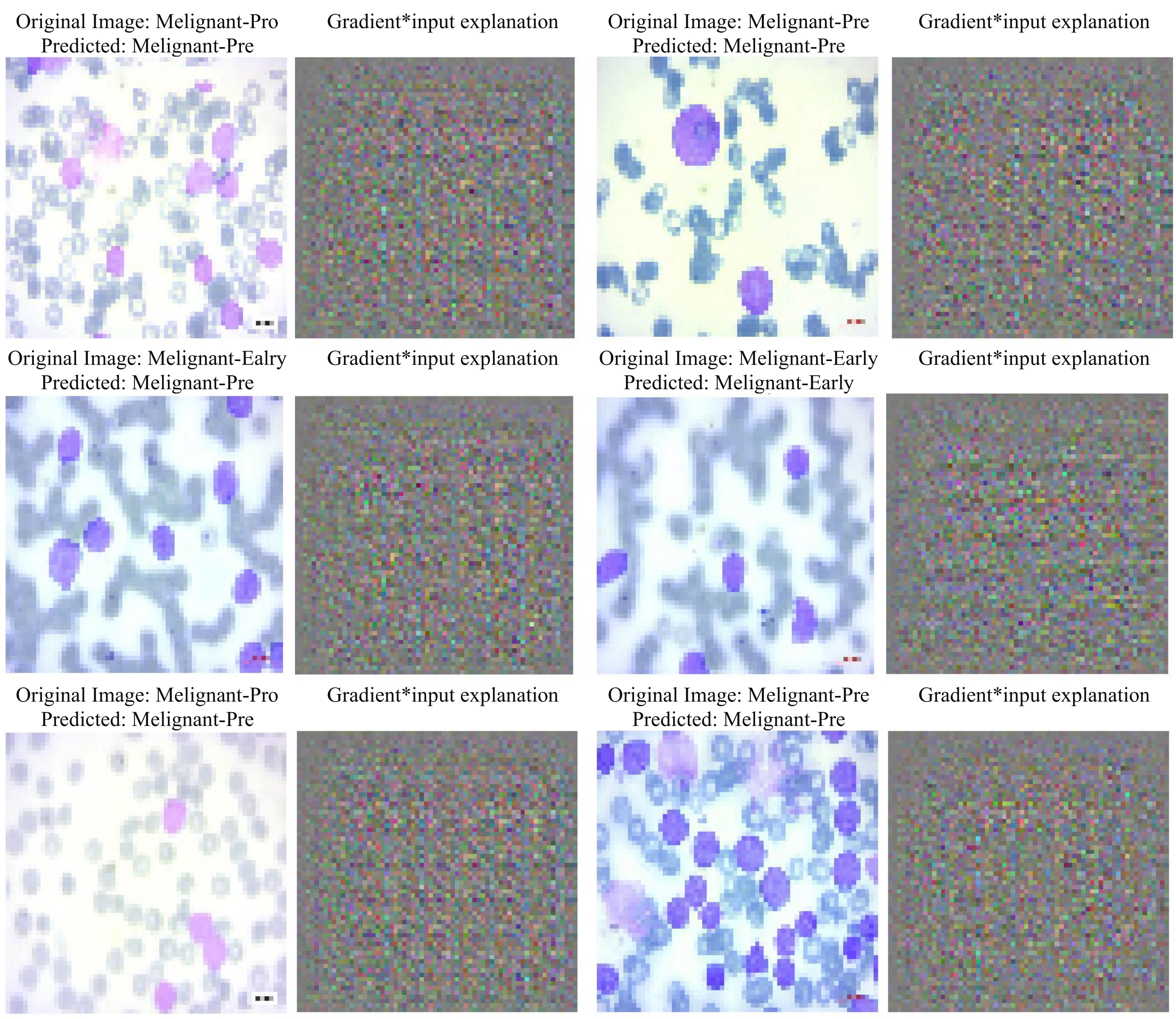

Figure 13: Pixel intensity: Heatmaps illustrating spatial attention for representative correct and incorrect classifications in PBS images.

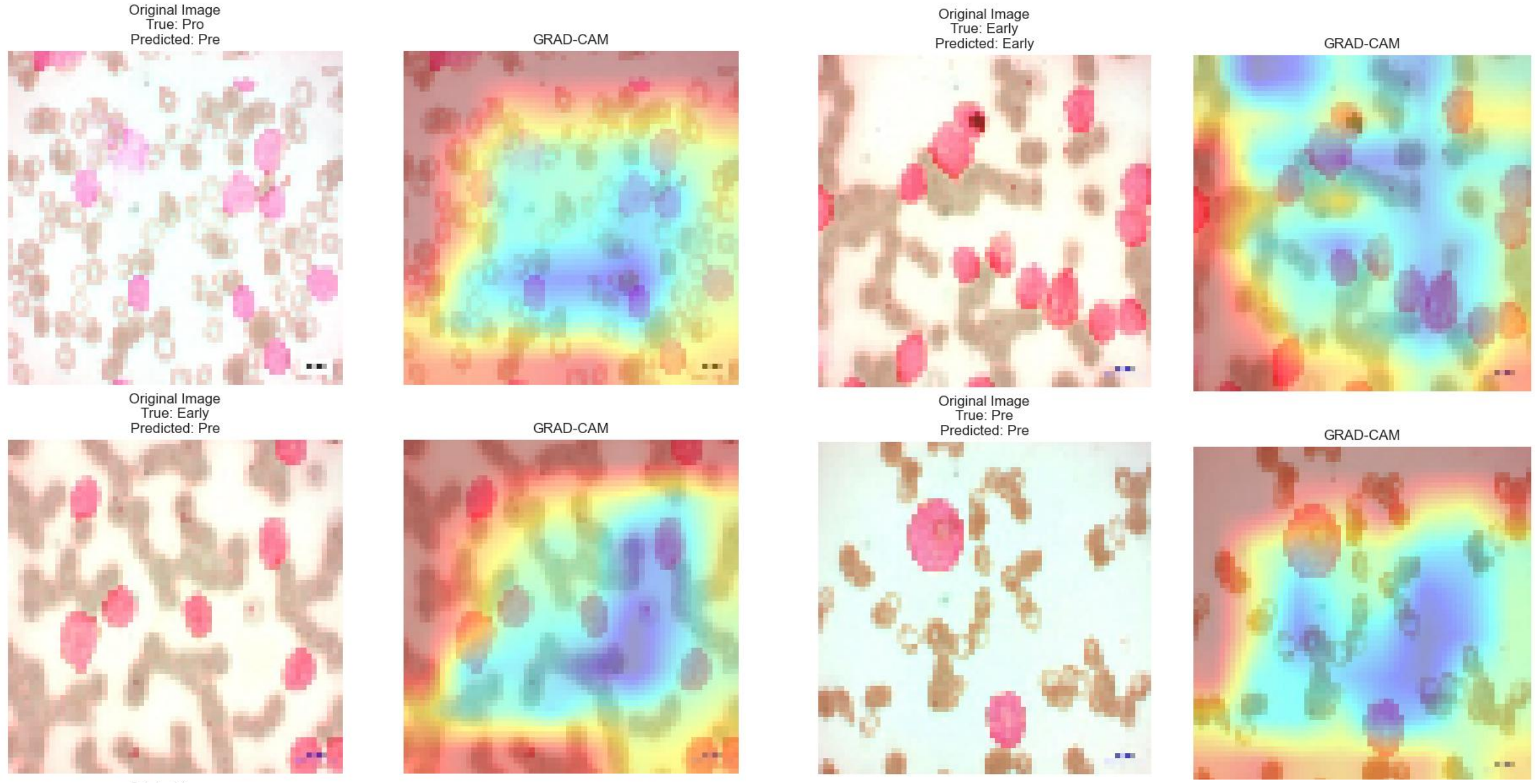

Figure 14: Grad-CAM view. Heatmaps illustrating spatial attention for representative correct and incorrect classifications in PBS images.

# 5. Concept of Smart acute lymphoblastic leukemia detection and management

Since leukemia patients require early diagnosis, real-time monitoring, and faster responses from medical practitioners, a sensor-based Smart Acute Lymphoblastic Leukemia Detection and Management (SLMS) system is proposed. In this proposed system, the following components are required: an IoT camera, an IoT Gateway, a machine learning server (MLS), and an IoT router (see Figure 21). The MLS serves as the central hub of the system. The trained ML model resides in the MLS. The system proposes using a wireless connection to connect all devices. Edge devices serve as delivery nodes in the IoT framework.

The devices are connected to the hospital database and communicate via secure protocols (e.g., Wi-Fi, Bluetooth, or Ethernet). Appropriate applications, such as web-based systems and mobile app software, are required for seamless integration with existing healthcare systems for data retrieval, processing, and display. A high-level view of the system is presented in Figure 26.

histopathology images are input into the system via IoT Cameras installed in the hospitals or handheld imaging devices. The system supports wireless data transmission from portable imaging devices, enabling real-time processing in resource-constrained settings. Doctors or authorized personnel can upload photos directly through a connected web interface or via external devices, such as USB drives.

The trained LukeNet model will reside in MLS. It will collect data from image processing and evaluate whether any diseased leaves are present. If it detects any disease on the histopathology image, it will propagate to the user interface. A secure IoT connector connects the MLS to users' edge devices, such as computers, laptops, or mobile phones.

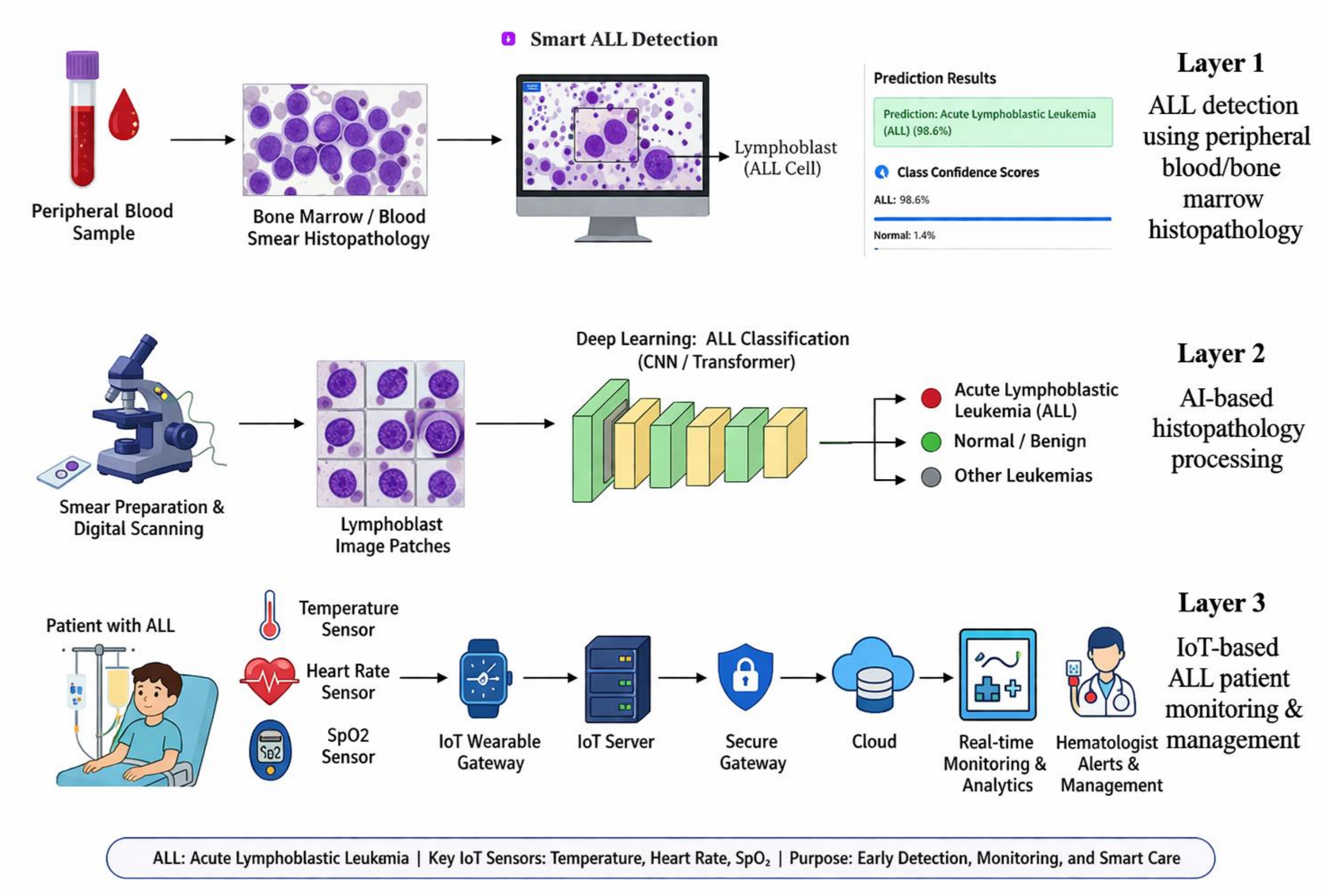

Figure 26: The proposed IoT-based leukemia detection and monitoring system

**Layer 1:** The MLS processing layer collects data directly from histopathology via Picture Archiving and Communication Systems (PACSs) or Radiology Information Systems (RISs). The histopathology images are processed via MLS computing and CNNs for histopathology detection and classification. The user interfaces are presented in Figures 27 and 28.

**Layer 2:** The histopathology detection and classification layer predicts histopathology, modalities, and associated probabilities. The results are displayed on IoT edge devices (e.g., a web dashboard or connected mobile application). This allows healthcare professionals to access actionable insights promptly. The leukemia class label and its confidence score are shown alongside the test image.

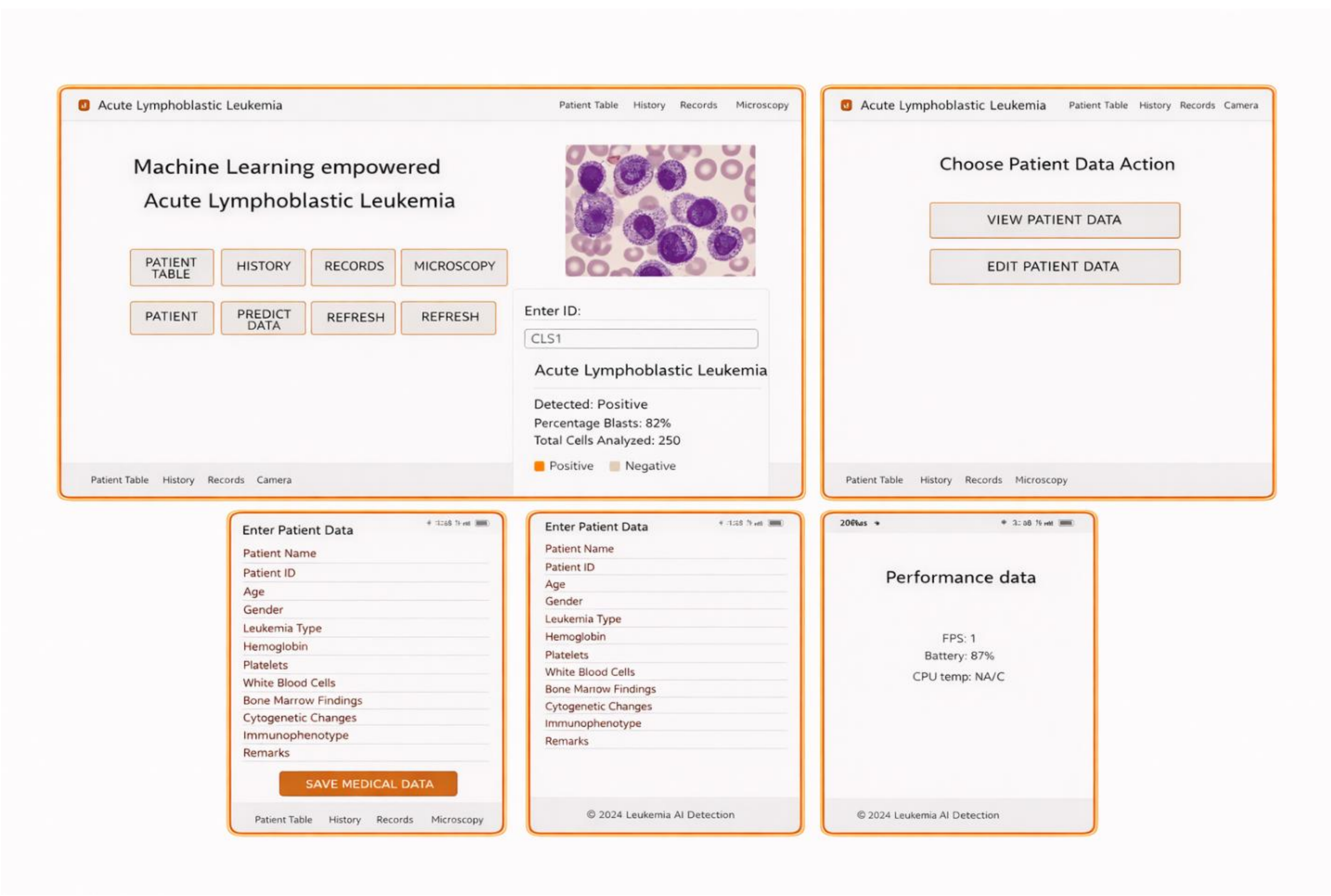

Figure 27: StreamLit web application interface and sample results

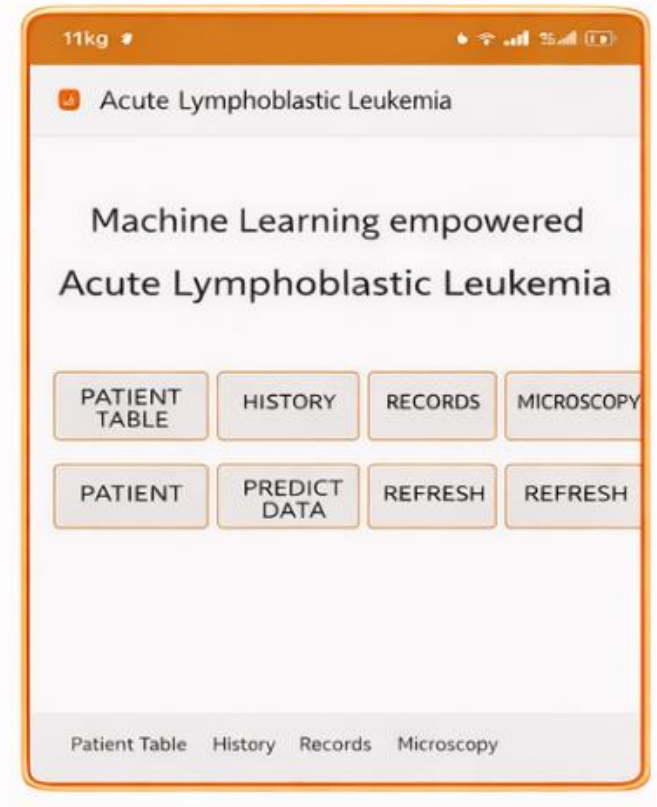
11kg
Acute Lymphoblastic Leukemia
Machine Learning empowered
Acute Lymphoblastic Leukemia
PATIENT TABLE
HISTORY
RECORDS
MICROSCOPY
PATIENT
PREDICT DATA
REFRESH
REFRESH
Patient Table
History
Records
Microscopy

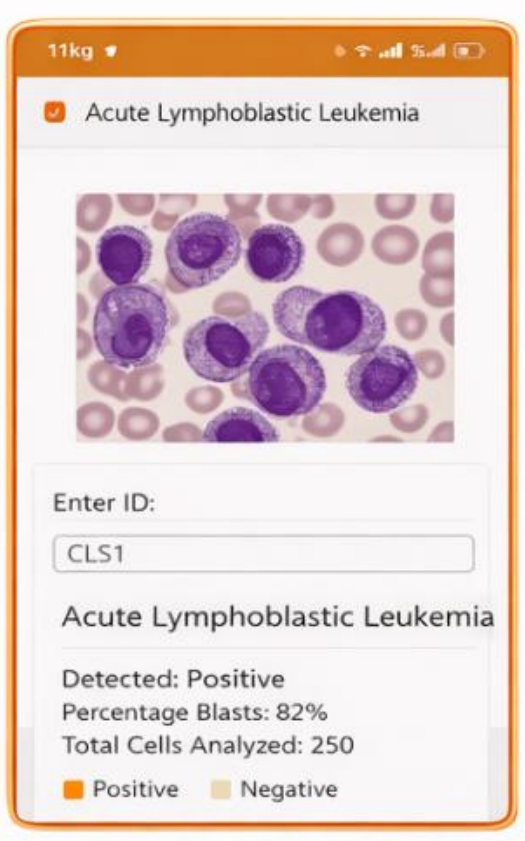
11kg
Acute Lymphoblastic Leukemia
Enter ID:
CLS1
Acute Lymphoblastic Leukemia
Detected: Positive
Percentage Blasts: 82%
Total Cells Analyzed: 250
Positive
Negative

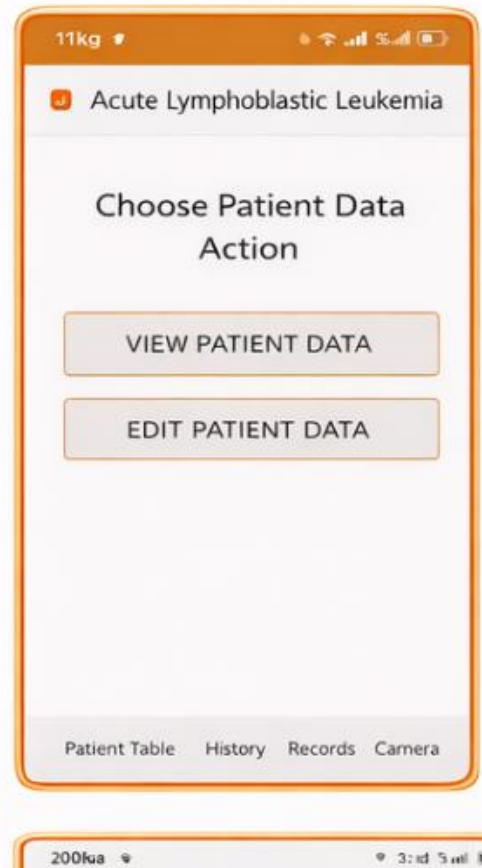
11kg
Acute Lymphoblastic Leukemia
Choose Patient Data Action
VIEW PATIENT DATA
EDIT PATIENT DATA
Patient Table
History
Records
Camera

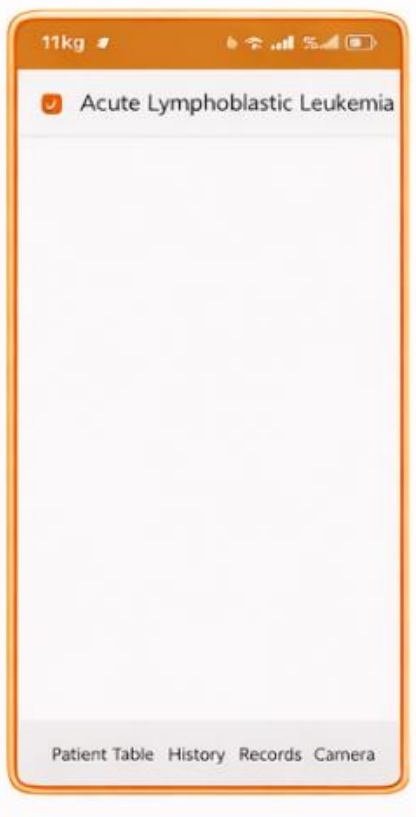
11kg
Acute Lymphoblastic Leukemia
Patient Table
History
Records
Camera

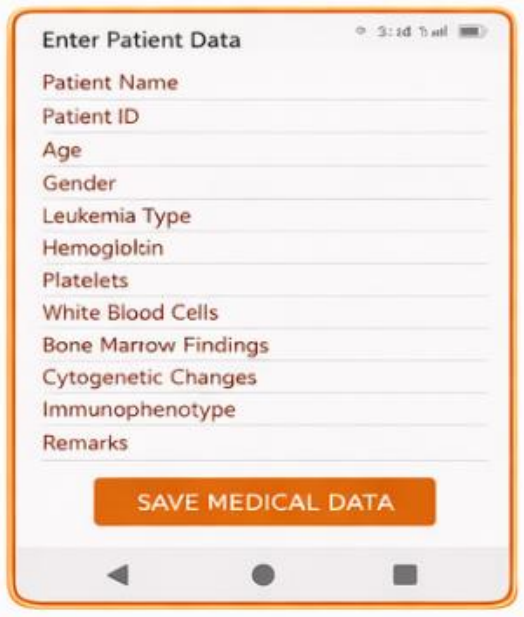
Enter Patient Data
Patient Name
Patient ID
Age
Gender
Leukemia Type
Hemoglobin
Platelets
White Blood Cells
Bone Marrow Findings
Cytogenetic Changes
Immunophenotype
Remarks
SAVE MEDICAL DATA

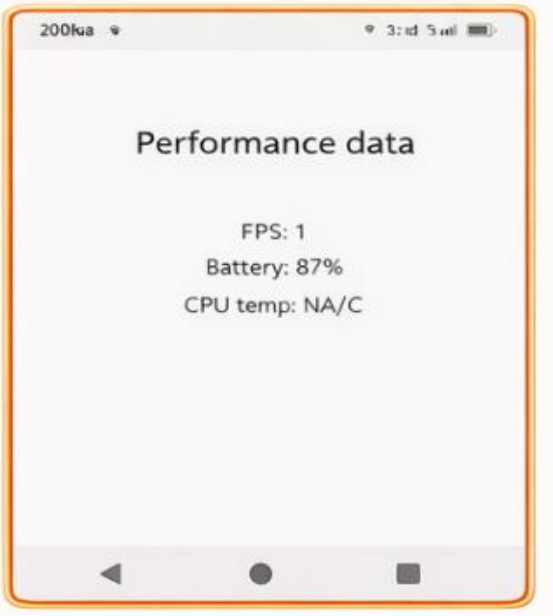
Performance data
FPS: 1
Battery: 87%
CPU temp: NA/C

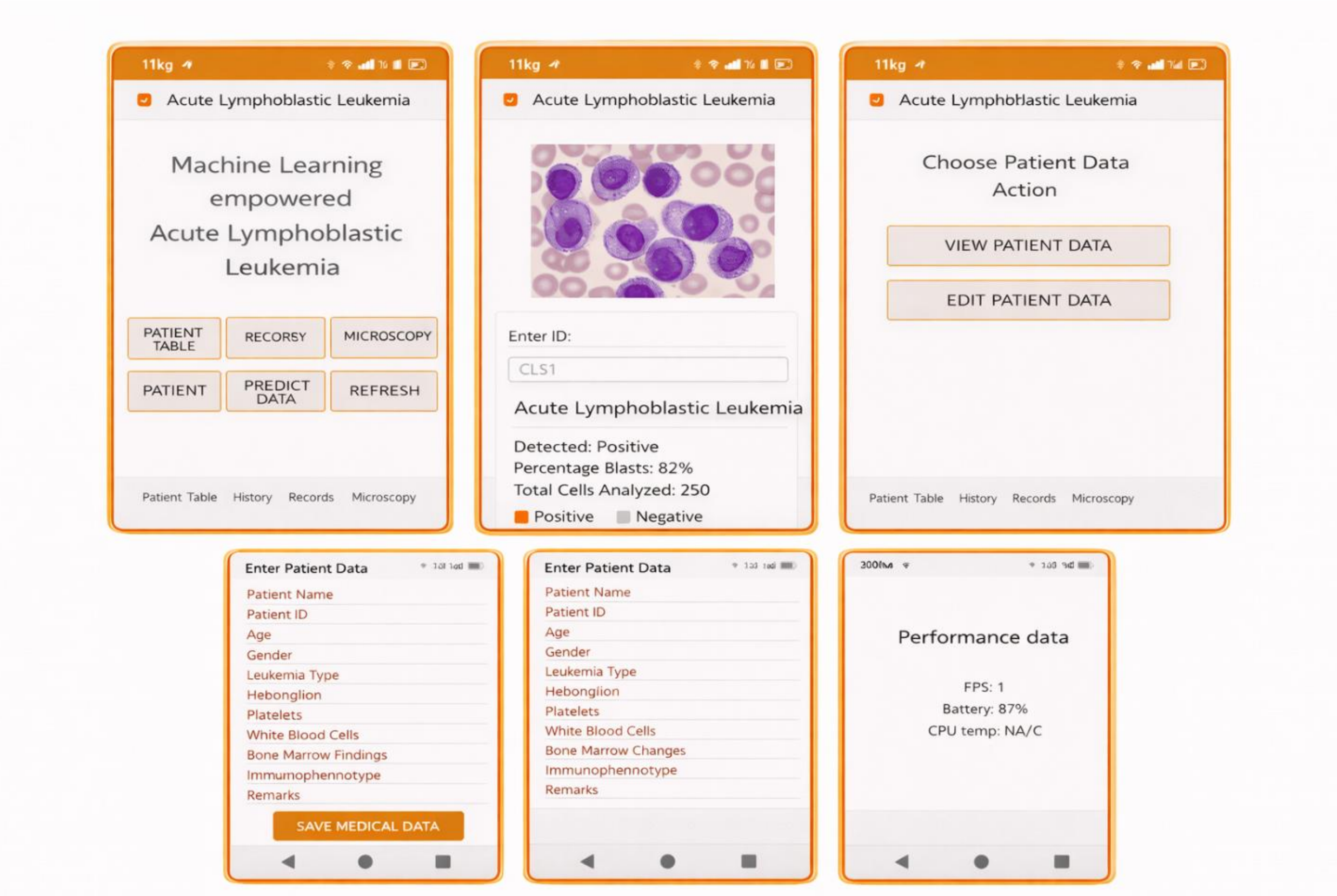

Figure 28: Android app interface integrated the LukeNet model

**Layer 3:** IoT-based monitoring provides continuous care, tracks symptoms, and improves the quality of life for leukemia patients. IoT devices can assist in providing continuous care, tracking symptoms, and improving the quality of life of leukemia patients. For example, wearable health monitors, seizure monitoring devices, brainwave monitoring systems, and remote patient monitoring devices enable continuous monitoring of leukemia symptoms, including changes in cognitive function, motor skills, and speech. They can transmit this data to healthcare providers.

The LukeNet application was tested on a Realme Note 70 mobile phone as an edge device. The smartphone features a 6300mAh battery, a 90Hz HD+ display, an IP54 dust- and water-resistant design, a Unisoc T7250 chip, and a 13MP primary camera. The LukeNet achieved a modest 1 FPS, indicating that the system processes one frame per second. The device's battery consumed 21% during operation. The temperature data showed that the CPU temperature did not increase significantly, as the data was recorded as "N/A." This suggests that the application did not impose a substantial thermal load on the device, even with continuous operation.

While the system has demonstrated functionality on edge devices, such as the Realme Note 70, we acknowledge that the inference speed of 1 FPS may not be sufficient for real-time clinical use. This performance was obtained under specific conditions and on hardware with moderate capabilities. The 1 FPS rate should be considered near real-time rather than real-time. We will revise the manuscript to emphasize that further optimization is needed to meet real-time processing requirements for clinical deployment. Optimizations such as model compression, quantization, or hardware acceleration on more powerful edge devices could significantly improve this performance.

The initial evaluation was conducted primarily on GPU hardware, which is not representative of mobile or edge-device conditions. Therefore, we will conduct additional CPU-only benchmarking on various mobile and edge devices to assess the model's performance in more realistic settings. Additionally, a memory footprint analysis will be included to evaluate the model's suitability for edge deployment by assessing its RAM consumption and computational demands on typical edge devices.

# 6. Discussions

Since the CNN model implemented with preprocessing for ALL detection and classification may exhibit biased accuracy in real-time implementation (Koresh, 2024; Obeagu, 2025), this study aimed to present a novel CNN that requires minimal preprocessing for ALL detection and classification. The CNN model, LukeNet, presented in this study, was applied to three datasets and achieved significant results. The LukeNet not only achieved high accuracy, but also performed well on unseen test data (99% accuracy). The results are better than the SOTA CNNs (DenseNet121, ResNet50, InceptionV3, Xception, MobileNet, VGG16) in terms of accuracy. This research investigated the performance of (i) six original CNN networks from spatial exploitation (VGG19), depth-based (ResNet152v2), multi-path (DenseNet201), width-based multi-connection (ResNext101), Feature-Map Exploitation (SE-ResNet152), (ii) six corresponding transfer learning, and (iii) two weighted ensemble model in ALL detection and classification of modalities of ALL datasets. The results are presented in Figure 16.

Among the original CNNs, DenseNet201 achieved the best accuracy at 98%. The study by Houssein et al. (2023) also used a deep DenseNet and achieved the best results for leukemia identification. The author purported that this occurs along the main path for information from the input to the output layer. The DenseNet connectivity pattern suggests that the network connects all layers directly. This process ensures that the information flow is maximum within the network. The feed-forward nature is maintained by obtaining inputs from the preceding layers for each layer. However, Kadry et al. (2022) and Wu et al. (2023) have already noted the challenge posed by gradient-based loss functions in hematology image analysis. These loss functions can also struggle with class imbalances common in medical datasets, where certain cell types are underrepresented. In such a scenario, DenseNet is a well-performing CNN model, with all layers connected.

Our investigation suggests that transfer learning with deep learning models achieves slightly higher accuracy than the original individual networks on small datasets (fewer than 2000 images). In this case, the accuracy was higher than the original CNN architecture only after careful training, including transfer learning. The transfer learning strategies in this research relied on using a pre-trained model for training and feature extraction. Surprisingly, we found that seresNext101 has improved by 17% accuracy after a transfer learning process. This is consistent with the study by Oloko-Oba and Viriri (2021), which found that SE-ResNeXt-101 typically requires more parameters and is computationally expensive, yet has shown promising results on ImageNet classification tasks. Performing transfer learning from images trained on Imagenet (general images such as cats, dogs, etc.) or MURA (X-ray images on different body parts but not the chest) improved results compared to scenarios when transfer learning was not used.

Not surprisingly, our investigation found that an ensemble of deep learning models improved accuracy compared to a single CNN architecture. Our findings also support the study by Acharya et al. (2020). The weighted ensemble model achieved 98.76% accuracy. This result provides better results than the study by Saeed et al. (2024), accuracy of 99.61%; Claro (2020), accuracy of 97.23%; Vogado et al. (2022), accuracy of 93%, with 95%; Sampathila (2022), accuracy of 95.54%; Cheuque et al. (2022), accuracy 98.4%; Khan (2021), accuracy 98%; Parayil et al. (2022), accuracy 89.75%, and Rastogi et al. (2022), accuracy 96.15%. However, studies by Ghaderzadeh et al. (2022), Rahman et al. (2023), Hosseini et al. (2023), and Veeraiah et al. (2023) have achieved better results than this study

# 7. Contributions

This study presents a novel CNN, LukeNet, that requires minimal preprocessing of PBS images. Although there are criticisms that CNNs' deeper layers receive little information from the input layer in classification tasks, LukeNet has achieved 99% to 100% accuracy across ALL detection tasks across three datasets. The properties of CNNs are balanced so that, even as the depth of the CNN increases, the gradient loss is minimized, and the model is trained correctly. The model was evaluated against six SOTA CNNs, transfer learning, and an ensemble model. In biomedical engineering, the consequences of misclassification are severe, and the accuracy of ALL detection and the classification of modalities are of concern, as lower detection accuracy and high false-positive rates will narrow the applicability and acceptability of CNNs in biomedical research (Hossain et al., 2023; Aladhadh et al., 2022). Therefore, Zhang et al. (2023) proposed conducting multiple experiments across varied datasets to understand the robustness and limitations of CNN models. In this line, Liao et al. (2023) argue for extensive experimentation across different architectures and datasets to draw reliable conclusions about model performance. Hendrickx et al. (2024) warned that using a single model should be avoided in many decision-support applications, where mistakes can have severe consequences.

This study conducts comprehensive experiments across ALL datasets rather than relying on a single model or method, thereby avoiding misleading results. It advocates conducting multiple experiments to ensure robust, reproducible findings. The following points in this study can be concluded as contributions:

1. The study presents a Deep CNN LukeNet that requires minimal image processing; however, it provides high accuracy. The model is appropriately positioned for clinical settings.

2. This study contributes to the ALL knowledge by finding that spatial exploitation (VGG19- accuracy 96.94%), depth-based (ResNet152v2- accuracy 96.99%), and multi-path (DenseNet201 – accuracy 98.08%) provide better performance than width-based multi-connection (ResNext101 – 86.41%), Feature-Map Exploitation (SE-ResNet152 – 90.93%)

3. This study pinpointed that negative transfer is caused by dissimilar target and source datasets. This outcome highlighted the effectiveness of transfer learning, though it can also successfully train deep learning models.

4. Even if a specific CNN architecture does not perform well, an ensemble of several models may still outperform individual models. Using DenseNet201, SE-ResNet152, and VGG19, this paper proposes an ensemble model (DVS) achieving the best accuracy of 98.76%. Saving the weights of individual CNNs in a .h5 file and then using them for testing takes less time than other ensembling approaches.

# 8. Limitations and Future Scopes

Like much other research, this research also has limitations and future scope for improvements. Firstly, one can argue that the model was tested on only one dataset. Hence, it is uncertain how the model will perform in other datasets. In the future, the model will be tested on more datasets than ALL. Secondly, the model was tested on PBS images; in the future, the approach can be tailored to mammography, histopathology, and MRA images. Lastly, this research has not yet been applied, but it presents a high-accuracy model. In the future, the model can be deployed in a machine that can be applied to CAD.

# 9. Conclusion

The growing global concern about ALL underscores the need for advanced diagnostic tools, such as computer-aided disease diagnosis, to improve early detection and patient outcomes. Traditional convolutional neural networks (CNNs), while effective, often require extensive preprocessing and limited adaptability across diverse imaging datasets, which pose challenges for widespread clinical adoption. To address these issues, this study proposes MI-NET, an interpretable CNN architecture explicitly designed for PBS classification of ALL. With a mere 11 million parameters, LukeNet optimizes computational efficiency while integrating post-hoc explainability tools, such as SHAP and Grad-CAM, to offer clear insights into its decision-making process. LukeNet achieved over 99% accuracy across three datasets, surpassing SOTA CNNs, transfer learning models, and ensemble models. Its strong discriminative power and adaptability to varied imaging conditions mark a notable advance over conventional methods, while its reduced parameter count makes it suitable for resource-limited healthcare settings. LukeNet is currently limited to grayscale mammograms and may require adaptation to a formulation or color imaging. Post-hoc explainability could be improved through inherently interpretable architectures. Accuracy is the most crucial element in computer-aided disease (CAD) diagnosis, and the LukeNet model has emerged as a practical approach for integrating into CAD. This is a step towards the application of artificial intelligence in biomedical engineering. This study will strengthen the CAD system, as it has proven more effective than humans in cancer diagnosis. Future work should explore multi-view or temporal mammography, integration with other modalities (e.g., ultrasound, histopathology), and real-time edge deployment, alongside clinical validation and collaboration with radiologists to support adoption in breast cancer screening.

# Data availability statement

The datasets used in this investigation are publicly available. The data for this research are stored in the Kaggle respiratory system.

https://www.kaggle.com/datasets/mohammadamireshraghi/blood-cell-cancer-all-4class

https://www.kaggle.com/datasets/mahdinavaei/blood-cancer

https://www.kaggle.com/datasets/sumithsingh/blood-cell-images-for-cancer-detection

# Funding statement

This research did not receive any specific grant from funding agencies in the public, commercial, or not-for-profit sectors.

# Credit authorship contribution statement

Md Taimur Ahad: Writing – original draft, project administration, methodology, investigation, formal analysis, conceptualization.

# Declaration of Competing Interest

The authors declare that they have no known competing financial interests or personal relationships that could have influenced the work reported in this paper. Acknowledgment: The authors would like to thank the anonymous reviewers for their valuable comments and suggestions.